\documentclass[acmsmall]{acmart}
\renewcommand\footnotetextcopyrightpermission[1]{} 
\usepackage[graphicx]{realboxes}

\usepackage{booktabs} 
\usepackage{rotating} 
\usepackage{multirow}
\usepackage{siunitx} 
\usepackage{adjustbox}
\usepackage{longtable}
\usepackage{paralist}
\usepackage{enumitem} 

\usepackage{graphicx}
\usepackage{pdflscape}
\usepackage{afterpage}
\usepackage{longtable}

\newcolumntype{C}[1]{>{\centering\let\newline\\\arraybackslash\hspace{0pt}}m{#1}}

\begin{document}
\sloppy
\title{Survey on Automated Short Answer Grading with Deep Learning: from Word Embeddings to Transformers}

\author{Stefan Haller}
\affiliation{%
  \institution{University of Twente}
  \city{Enschede}
   \country{The Netherlands}
}
\email{s.m.haller@student.utwente.nl}

\author{Adina Aldea}
\affiliation{%
  \institution{University of Twente}
  \city{Enschede}
  \country{The Netherlands}
}
\email{a.aldea@utwente.nl}

\author{Christin Seifert}
\affiliation{%
  \institution{University of Twente}
   \city{Enschede}
   \country{The Netherlands}
   \institution{ - University of Duisburg-Essen}
   \city{Essen}
   \country{Germany}
}
\email{christin.seifert@uni-due.de}

\author{Nicola Strisciuglio}
\affiliation{%
  \institution{University of Twente}
   \city{Enschede}
  \country{The Netherlands}
}
\email{n.strisciuglio@utwente.nl}

\begin{abstract}
Automated short answer grading (ASAG) has gained attention in education as a means to scale educational tasks to the growing number of students. Recent progress in Natural Language Processing and Machine Learning has largely influenced the field of ASAG, of which we survey the recent research advancements. We complement previous surveys by providing a comprehensive analysis of recently published methods that deploy deep learning approaches. In particular, we focus our analysis on the transition from hand-engineered features to \emph{representation learning} approaches, which learn representative features for the task at hand automatically from large corpora of data. We structure our analysis of deep learning methods along three categories: word embeddings, sequential models, and attention-based methods. Deep learning impacted ASAG differently than other fields of NLP, as we noticed that the learned representations alone do not contribute to achieve the best results, but they rather show to work in a complementary way with hand-engineered features. The best performance are indeed achieved by methods that combine the carefully hand-engineered features with the power of the semantic descriptions provided by the latest models, like transformers architectures. We identify challenges and provide an outlook on research direction that can be addressed in the future.
\end{abstract}

\maketitle

\section{Introduction}
\label{sec:intro}

In the education field, large attention is dedicated to characterizing the learning process of students to determine the efficiency and success of learners in acquiring new knowledge. 
Assessing and quantifying the knowledge gain are fundamental aspects for evaluating the quality of the learning process. Due to the time-consuming and individual nature of the assessment process, it is crucial to render it as efficient as possible, e.g. by using some level of grading automation, while retaining or even improving its quality~\cite{Mohler2009}.

Besides formative assessment (e.g. teacher feedback of assignments), the main method of assessment is via summative assessment (e.g. written examinations)~\cite{Suzen2018}, where knowledge acquired by the students can be tested in several ways. Different types of questions can be formulated: closed-form questions (e.g. multiple-choice) or open answer questions (e.g. essays or short answers)~\cite{Burrows2014}. Automated grading for multiple-choice questions is straightforward and immediate.
Automated Essay Scoring (AES) and Automated Short Answer Grading (ASAG) are more challenging tasks because the assessment of these answers requires an understanding of the text and a more detailed analysis~\cite{Magooda2016,Suzen2018}. The main difference between AES and ASAG is that the latter deals with short answers usually graded against a reference answer, whereas the former is concerned with scoring longer textual answers and the grading is based on evaluating the quality in terms of spelling, grammar and coherence, and less on compact information content~\cite{Magooda2016}.


The assessment of textual open answers is challenging and requires techniques and approaches from different fields, such as natural language processing (NLP), text understanding, and reading comprehension. Recent progress in Deep Learning and NLP facilitated the design and analysis of machine learning models for textual data~\cite{Suzen2018,Neterer2018DeepLI} and raised the question of their applicability to different application areas, such as AES and ASAG. 
These advancements and their application to ASAG have not been covered in recent surveys, which only reviewed methods and approaches based on feature extraction used in combination with classical machine learning models~\cite{Burrows2014,Galhardi2018MachineLA,Blessing2019AML}. This research focuses solely on ASAG, for which it is fundamental to construct an effective semantic representation of the answers.

In this paper, we extend previous surveys by analyzing recently published methods for ASAG based on deep learning. We first provide a historical perspective with an overview of the changes and developments in the field over time and subsequently dive deeper into the contributions of the most recent methods. We pay attention to the role of text representations, i.e. features, their importance to effectively describe the characteristics of sentences and paragraphs, and the shift from hand-engineered to automatically learned features determined by the use of deep learning methodologies. We organize the works following a simple taxonomy that focuses on the type of text representation used. It includes methods based on \emph{classical machine learning} , which combine hand-engineered features with classifiers, and \emph{deep learning} methods that are able to learn relevant features directly from training data. We group the methods in the first category according to the type of features, namely lexical, syntactic, and semantic features. The methods in the second group, deep learning-based methods, are organized according to the mechanisms used to learn textual representation, such as word-embeddings, sequential models, and attention-based models.
Further, we identify trends in the design of the network architectures and provide an outlook on future developments and challenges that need to be addressed, such as improving the textual understanding in cross-language settings and the generalization of actual methods to work with cross-domain texts (e.g. for different disciplines). 

The remainder of the article is organized as follows. In Section~\ref{sec:contributions} we discuss the contributions of the present work with respect to previous survey approaches, while in  Section~\ref{sec:historical} we provide a view of the historical developments related to the research field of ASAG. In Section~\ref{sec:organization}, we elaborate on the scope and organization of the survey, while in Section~\ref{sec:datasets}, we present the details of the most used benchmark data sets. In Section~\ref{sec:taxonomy}, we provide a taxonomy for the organization of the existing works. In Section~\ref{sec:ML} and Section~\ref{sec:DL}, we analyze the works based on classical machine learning and deep learning approaches, respectively. We discuss actual trends and provide an outlook for future developments in the field of ASAG in Section~\ref{sec:discussion}, and draw conclusions in Section~\ref{sec:conclusions}.

\section{Contributions}
\label{sec:contributions}

The developments and progress in the field of automated short answer grading (ASAG) have been previously reviewed in few survey papers~\cite{Burrows2014,Galhardi2018MachineLA,Blessing2019AML}. A unified and comprehensive overview of the entire field can be found in~\cite{Burrows2014}, where the developments of analytic components until 2014 were addressed from a historical perspective. The authors organized the existing approaches into five groups, which correspond to temporal themes, also called \emph{eras}. 
Methods categorized as belonging to a certain era share the approach and focus: word and sentence matching, feature extraction and comparison, use of semantic information, inference based on machine learning tools, and word embeddings. 

A comprehensive analysis of approaches that combined feature engineering with machine learning-based predictive models was presented in~\cite{Galhardi2018MachineLA}. The authors provided a systematic literature review focusing on existing data sets, applied NLP techniques, and machine learning algorithms for ASAG. They highlighted the role and importance of a proper feature set design for effective training of machine learning models. Other published reviews~\cite{Blessing2019AML} did not discuss further aspects of the developments in the field of ASAG. 

More recent advancements in short answer grading with deep learning methods have not been covered yet, although they concern major progress and an increase of performance. In this work, we address this gap and aim at complementing previous surveys with a thorough analysis and organization of methods based on deep networks and representation learning techniques. 
We render the details of the transition from hand-engineered features to semantic-rich text representations as the key driver for a substantial improvement of performance on ASAG and NLP tasks. 
We trace the deep learning developments of ASAG systems back to three stages, related to the type of techniques used to learn textual representations: (1) models based on word embeddings, (2) sequence-based models that learn entire sentence representations, and (3) attention-based models. With this organization of existing models, the present survey enables future extensions while providing a comprehensive and structured overview of published work. 
The key contributions of this survey are:
\begin{itemize}[topsep=0pt, partopsep=0pt]
    \item we provide a comprehensive comparison of the latest methods for automated short answer grading and their performance on different data sets; 
    \item we present an overview of the main benchmark data sets for evaluation of ASAG systems;
    \item we identify the actual trends and most promising model architecture for advancing performance in ASAG; 
    \item we present an analysis of the impact of progress in NLP and deep learning methods on the field of ASAG.
\end{itemize}

\section{Historical perspective}
\label{sec:historical}

The development of the ASAG field has gone through a number of milestones. In~\cite{Burrows2014}, the authors identified some trends of the progress  and associated them with specific time periods. Initially, researchers used \textbf{concept mapping} to compare answers to a reference answer by breaking them down into several concepts. The inferred concepts were used as basic reference models. Predictions of the correctness of the answers were made by inspecting their similarities with the reference answerss.
These approaches were extended or partly replaced by methods developed in the area of \textbf{information retrieval}, 
which were more focused on the extraction of relevant characteristics, i.e., features, from the student answers for direct comparison with reference answers. Semantic information and relations between words in sentences were not taken into account during these initial stages of research. These mainly concerned analyses of text, based on regular expressions or parse trees~\cite{Burrows2014}. 
Subsequently, \textbf{semantic similarities between sentences} were modeled and exploited by using large corpora of text, e.g. WordNet ~\cite{Fellbaum2000WordNetA}, word synonyms, and the development of knowledge-based methods. Semantic features improved the overall performance of ASAG methods and made them more flexible ~\cite{Magooda2016}. 

The development of Machine Learning methods to construct predictive models also influenced the field of ASAG. The inference capabilities of trainable classification systems shifted the focus of researchers towards studying the effect of using different sets of features for the extraction of relevant characteristics of the text, namely \textbf{lexical}, \textbf{syntactic}, \textbf{semantic features}, and combinations of them. 
Extensive feature engineering was needed to construct reasonable sets of features that could perform well on a given problem and on data sets with specific characteristics. The construction of feature sets thus required domain knowledge by human experts, which could guarantee the design of systems that achieved good performance results on specific data sets~\cite{Ott2013CoMeTID}. Papers that proposed methods in this category demonstrated that semantic feature extraction methods were key to obtain high performance.

A major milestone in the field of NLP was the development of \textbf{word embeddings} methods, which consist of techniques for mapping words or phrases from a vocabulary into a vector space with interesting and useful properties for predictive tasks. 
The aim of word embeddings is to focus on the properties those feature spaces are supposed to retain, e.g. capturing semantic similarity. This had a significant impact on the field of ASAG, as it improved the capabilities of machine learning-based systems to extract meaningful semantic information from text, compared to previous methods based on hand-crafted features. 
The development of \textbf{sequential machine learning models}, e.g. Recurrent Neural Networks (RNNs)~\cite{DEMULDER201561,dyer-etal-2016-recurrent} and Long Short-Term Memory networks (LSTMs)~\cite{Cheng2016}, able to learn dependencies in sequences of words (sentences and paragraphs) contributed to substantial improvements of the computed text representations. 
Long-range dependencies in text were then successfully modeled via \textbf{attention-based neural networks}, first combined with recurrent networks~\cite{KimDHR17} and then as stand-alone models~\cite{Vaswani2017AttentionIA}. Attention-based approaches were applied in tasks like detection of semantic textual similarity~\cite{Lan2020ALBERTAL,Liu2019RoBERTaAR,Raffel2019ExploringTL}, paraphrase identification~\cite{Yang2019XLNetGA,Liu2019ImprovingMD,Ratner2019TrainingCM}, reading comprehension~\cite{Lan2020ALBERTAL,Raffel2019ExploringTL,Zhang2019SemanticsawareBF,Devlin2019BERTPO}, and recognizing textual entailment~\cite{Liu2019RoBERTaAR,Yang2019XLNetGA,Liu2019ImprovingMD}. Methods that model long-range dependencies in text became the state-of-the-art approaches for most NLP and ASAG tasks.


\section{Organization of the review}
\label{sec:organization}

The survey is structured in three parts. The first part  covers important benchmark data sets for the evaluation of short answer grading (ASAG) algorithms (cf. Section~\ref{sec:datasets}). Each of the data sets has specific characteristics that allow to evaluate different aspects of ASAG systems related to their generalization properties, such as grading of answers to questions unseen during the training phase or about new topics or domains. 

We track the progress and performance improvements of methods based on classical machine learning tools and hand-engineered feature sets, and on more recent deep learning approaches in the second and third part of the survey (cf. Sections~\ref{sec:ML} and~\ref{sec:DL}), respectively. We pay particular attention to their capabilities to extract semantic-rich text representations. We analyze recently published methods and elaborate on the latest developments and trends in the field of ASAG.

\subsection{Scope}

We focus on the analysis of methods for automated grading of short answers, defined by the following criteria~\cite{Burrows2014}:

\begin{enumerate}
    \item the answer has to reflect the student knowledge and should not report just passages from a provided prompt text; 
    \item the answer is given in natural language; 
    \item the answer's length is around $50$ words, but can contain up to about $100$\footnote{Some of the analyzed datasets contain outliers (see table~\ref{tab:datasets}) that exceed that threshold but were nevertheless taken into account, as the majority of responses were below}; 
    \item grading of the answer is focused on the content rather than the writing quality; 
    \item the possible answers are restricted by the closed-form of the question.
\end{enumerate}

Most of the approaches for ASAG published in recent years were based on supervised learning strategies and designed to support grading efficiency in educational settings~\cite{Burrows2014,Galhardi2018MachineLA}. Unsupervised approaches, such as ranking or clustering, were used to group together student answers based on similarity, aiming at improving the uniformity of grading and can be used as a complementary tool to supervised learning methods.

In practice, in the case of courses whose content does not change drastically over time, the question pool used for examinations is slightly expanded or does not significantly change, and the same questions are re-used. This results in the availability of multiple answers to the same questions, which can be used as training inputs for the life-long optimization of the models. In this context, with the availability of labeled questions, reference answers, and student answers, automated grading systems based on supervised learning are mostly investigated. 
Hence, we focus on reviewing supervised methods, which are largely deployed for ASAG, and provide more precise evaluations due to the use of labeled correct answers for training.

\subsection{Search methodology}
We applied a semi-systematic approach for the literature review, which consisted of the following steps: 
\begin{inparaenum}[\itshape a\upshape)]
\item keyword-based search in scientific databases;
\item focus on benchmark data sets and methods tested on them; 
\item backward search based on references of relevant papers.
\end{inparaenum}

We searched for relevant papers in the main scientific databases, namely ScienceDirect, Google Scholar, ResearchGate, Semantic Scholar, and arXiv. The main search terms that we used were `automatic grading system', `automated assessment', `digital assessment of students', `short answer grading', and adapted them depending on the results. We subsequently identified the most used benchmark data sets (see Section~\ref{sec:datasets}), and selected the papers that report results on them. This allowed to include in the review those papers that propose systems for automated grading, also in the case they present them using slightly different task definitions, such as inference on answer similarity or automatic question answering.
Finally, we gathered further papers by investigating the reference lists of the already selected ones. The criteria for the selection were the number of citations and year of publication, agreement with the topic concerned and application, and testing on similar data sets.


We especially focused on the papers that were published in the past five years, in order to identify the most recent developments and trends. The increasing number of published papers 
in the past years indicates a growing interest in ASAG systems and in their use for educational purposes.


\section{Benchmark data sets for Short Answer Grading}
\label{sec:datasets}

Existing methods have been tested on different data sets, e.g. SciEntsBank, Beetle, Texas, ASAP-SAS, and various others. These data sets differ in many aspects: number of questions, number of answers, question type, domain, language, grading basis, and answer length. For our analysis, we rely on the results reported in the original papers of the methods that we surveyed. They were tested usually on sub-sets of the available data sets, and a direct comparison of the performance is sometimes not possible. This is mainly due to the different composition and characteristics of the available public data sets. 
Some data sets are proprietary or have an unknown source, thus we exclude them from our analysis (e.g. the Large-Scale Industry Dataset~\cite{LSdataset2018,Saha2019JointML}). 

In this review, we focus on the four most widely used data sets for benchmarking ASAG methods: SciEntsBank~\cite{Dzikovska2013SemEval2013T7}, Beetle~\cite{Dzikovska2013SemEval2013T7}, Texas~\cite{Mohler2011LearningTG}, and ASAP-SAS~\cite{asap-sas} data sets. Their public availability and  diversity of the answer domains allow to evaluate different aspects of the performance and capabilities of automated grading algorithms. Furthermore, they guarantee a fair comparison of existing methods. In the following section, we describe the data sets and provide information on the specific task and applications for which they were designed. In Table~\ref{tab:datasets}, we summarize the characteristics and details of the composition of the considered data sets.

\begin{table*}
  \caption{Detailed characteristics of the SciEntsBank, Beetle, Texas2011 and ASAP-SAS benchmark  data sets. "Additional information" indicates whether additional textual information for the task besides the question itself is available.}
  \label{tab:datasets}
  \Small
  \renewcommand{\arraystretch}{1.2}
  \begin{tabular}{ p{4cm}C{2cm}C{2cm}C{2cm}C{2cm}  }
    \toprule
    \bfseries Characteristics & \bfseries SciEntsBank & \bfseries Beetle & \bfseries Texas2011 & \bfseries ASAP-SAS\\
    \midrule
    Training question/answer pairs & 4,969 & 17,198 & 2,442 & 17,207\\
    \midrule
    Percentage of correct answers & 40.41\% & 42.49\% & 44,22\% & 21.57\%\\
    \midrule
    Number of domains & 12 & 2 & 1 & 4\\
    \midrule
    Number of questions & 135 & 47 & 85 & 10\\
    \midrule
    Average number of answers per question & 37 & 366 & 29 & 1,721\\
    \midrule
    Average answer length (in words) & 13 & 10 & 18 & 42\\
    \midrule
    Maximum answer length (in words) & 110 & 80 & 173 & 325\\
    \midrule
    Minimum answer length (in words) & 1 & 1 & 1 & 1\\
    \midrule
    Additional information & No & No & No & Yes\\
    \midrule
    Scale of labels & 2-way, 3-way, 5-way \newline classification  & 2-way, 3-way, 5-way classification  & Score between 0 and 5 & Score between 0 and 2 or between 0 and 3\\
    \midrule
    Publicly available  & Yes & Yes & Yes & Yes\\
    \bottomrule
  \end{tabular}
\end{table*}

\subsection{SciEntsBank and Beetle data sets}

The SciEntsBank and Beetle data sets are part of the SemEval 2013 challenge~\cite{Dzikovska2013SemEval2013T7}, which has the objective of identifying common mistakes, such as omissions and wrong or thematically irrelevant statements, in order to develop customized correction strategies. The challenge is structured to evaluate different aspects of short answer grading systems.

The data sets include three sets of labels, which serve to train models on 2-,3- and 5-way task problems. The  2-way task consists of classifying the answers either as correct or incorrect, whereas for the 3-way task each answer is labeled as either correct, contradictory or incorrect. The 2- and 3-way tasks concern the Recognizing Textual Entailment (RTE) task.
The 5-way task aims at evaluating the identification of non-domain, correct, partially correct incomplete, contradictory and irrelevant answers. It is aimed at improving dialogue systems for tutoring.

The test protocol is designed to evaluate the performance of the algorithms on three sets of data, which represent possible challenging situations where the grading systems can be used, namely: 
\begin{inparaenum}[\itshape a\upshape)]
\item unseen answers during the training phase, 
\item unseen questions not used to train the models, and
\item questions/answers from unseen domains.
\end{inparaenum}  
These tests are meant to evaluate different aspects of the generalization properties of automated grading methods and their applicability to questions and domains other than those they are trained for.

\paragraph{\bfseries SciEntsBank}  
The SciEntsBank data set comprises student answers to questions collected as part of a standardized assessment in grades $3$ to $6$ in schools across North America. In total, the data set contains $4969$ answers to $135$  questions from $12$ domains (see Table~\ref{tab:datasets} for details). Depending on the domain category, the required grading concerns either a 2-way, 3-way, or 5-way classification.

\paragraph{\bfseries Beetle}
In contrast to the SciEntsBank data set, the Beetle data set is designed to test the interaction of students with a real tutorial dialogue system. 
The system teaches students in high-school physics and fundamentals of electricity and electronics. In order to create the data set, the dialogues have been revised and only relevant answers (not the interaction protocol) to questions were used. Questions are either factual questions, or explanation and definition questions. The corpus contains in total 47 questions with on average 366 student answers per question. The Beetle set only contains the categories of unseen answers and unseen questions. 

\subsection{University of North Texas data set}
The University of North Texas data set (Texas2011) is also widely used to evaluate and compare the performance of  automated grading systems. It contains answers given by 30 students to 80 different questions, which result in approximately 2400 question-answer pairs. The average answer length is 50 word tokens. The questions are collected from 10 assignments of two examinations of basic knowledge in the field of Computer Science. The answers are provided with a grade given by two assessors, which ranges between 0 and 5. The grades are given as integer numbers. There were no clear rules in the grading process and the average grade between the two assessors is considered the ground truth~\cite{Mohler2011LearningTG}. 
The data set has the purpose of mirroring real-world issues of the process of grading short answers as realistically as possible. This results in more complex and nested answer structures, which also include the use of tables.

\subsection{ASAP-SAS data set}
The Automated Student Assessment Prize Short Answer Scoring (ASAP-SAS) data set was released as part of a Kaggle competition in 2013, sponsored by the Hewlett Foundation~\cite{asap-sas}. It consists of ten questions from different domains, e.g., English, Biology, English Language Art, Science. In total the training data set contains $17,207$ answers (about $1,700$ per question) and the test set contains $5,224$ answers. On average an answer has $50$ words, although a small amount of answers ($<5\%$) also contains more than 100 words. Each of the questions is marked with a score in the range of $0$ to $2$ or $0$ to $3$.

The competition and the data set were designed to challenge assessment systems with realistic and diversified questions. Due to the diversity of the ten questions, it further aims to support the development of scoring systems and their real-world application in the educational sector. One unique characteristic of this data set is that the question structure varies a lot: for instance, some questions are formulated to ask for specific information contained in a prompt text of 1 or 2 pages. Other questions also contain pictures or graphs and the student is requested to state her own observations and interpretation of the results.


\section{Taxonomy}
\label{sec:taxonomy}
\begin{figure}[!t]
    \centering
    \input{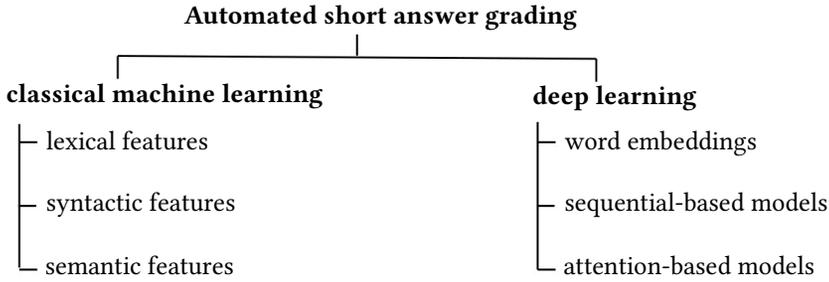}
    \caption{Taxonomy of Automated Short Answer Grading methods. The categorization of methods is based on the machine learning approach and on the type of representation (features) used. For classical machine learning approaches, the features are hand-engineered by domain experts, while in deep learning approaches the different types of representations are learned directly from the data. }
    \label{fig:taxonomy}
\end{figure}

Existing ASAG approaches fall into two broad categories, namely (1) early approaches which rely on hand-crafted features and classical machine learning (CML) approaches, e.g. logistic regression or support vector machines, and (2) deep learning (DL) approaches that formulate the feature design as a learning problem and combine it with training a predictive model~\cite{Sung2019ImprovingSA}. The second category of methods can be further divided into three sub-categories, which correspond to the phases of the development of NLP methods, namely word embedding, sequential models, and attention-based models~\cite{Young2018,Otter2020}. Figure~\ref{fig:taxonomy} shows an overview of the taxonomy of methods for the ASAG method. 

Word embedding techniques, most prominently Word2Vec~\cite{Mikolov2013EfficientEO}, aim to represent semantically similar words with vectors that are close in a learned latent space. These automatically learned representations are able to characterize  the information in large corpora of text effectively. Longer-range relations, in groups of words and longer sentences, were taken into account for the development of the second category of methods, which include approaches based on Recurrent Neural Networks (RNNs) and Long Short-Term Memory networks (LSTMs). These models capture relations between words of a sentence over longer distances and therefore provide richer sentence and paragraph representations. The models are usually trained for specific downstream tasks and can use the embedding results from methods such as Word2Vec~\cite{Sutskever14}. However, due to the vanishing gradient problem, recurrent networks tend to focus more towards short-term context information, while not being able to robustly catch longer-range dependencies in sentences. LSTMs alleviated this issue to some extent. The  limitation was finally addressed by neural network models based on attention mechanisms, which constitute the third category of deep learning-based methods that we identified. Attention-based models relax the strict sequential analysis of tokens in sentences and are able to model word dependencies without regard to their distance in
the sentences, also allowing for parallel processing of longer text sequences by means of multi-head attention mechanisms~\cite{Galassi2020}. Architectures fully based on attention are known as Transformers~\cite{Vaswani2017AttentionIA}.

\section{Hand-engineered features and Machine Learning}
\label{sec:ML}

Approaches based on hand-engineered features rely on the extraction of lexical, syntactic, and semantic features from the text, using dependency and constituency parsers, or  functions to compute the sentence overlap. 
The main goal of these features is to describe key components (e.g. specific terms, concepts) of good answers by detecting specific patterns. 

In most of the reviewed papers, the feature vectors extracted from the raw text were used in combination with a classical machine learning classifier, such as Logistic Regressor, Support Vector Machine, Random Forest, or Na\"{i}ve Bayes classifiers. Some of the reviewed approaches were based on ensemble methods combining predictions of several classifiers. 
In the following, we revise these approaches, organized in sections according to the type of features used to represent the text, namely 1) lexical, 2) syntactic and 3) semantic features, and elaborate on their relevance and use in existing methods. It is important to note that few approaches deployed features sets that were made of combinations of different types of features. This makes it difficult to draw absolute conclusions regarding the superiority of a certain feature or group of features over another. We provide an overview of the reviewed methods in Table~\ref{tab:CAT1.features} and the results that they achieved on benchmark data sets in Table~\ref{tab:CAT1.results}.


\subsection{Lexical features}
Lexical features have been largely used for the description of the characteristics of short textual answers. They consist of, for example, single words, stemmed or lemmatized words, their prefix or suffix. 
The extraction of lexical features is a simple process and several algorithms were proposed that use them in ASAG tasks, e.g. methods to compute the degree of overlap between sentences, n-gram representations, or lexical statistics.

Algorithms for estimating word overlap were a fundamental component of early ASAG systems. 
Overlap-based features measure the coinciding words between sentences, i.e. the student and reference answer, using methods to estimate the word-wise or character-wise overlap of two or more sample sentences. 
Often these methods were combined with different pre-processing methods, such as  lemmatization or stemming to further improve the quality of the results~\cite{Ott2013CoMeTID,Meurers2011IntegratingPA,Mohler2011LearningTG}. 
In ~\cite{Pribadi2017AutomaticSA}, the authors compared three-word overlap computation methods: i) Dice coefficient, ii) Jaccard coefficient and iii) cosine coefficient, and studied their impact on the performance of an ASAG system. These methods count the overlapping words in two sentences and compute a score of the similarity between them. The authors found that the cosine coefficient contributed to obtaining the best performance on the estimation of sentence similarity. In~\cite{Pribadi2016AutomatedSA}, a weighted cosine coefficient was used, resulting in a further improvement of the performance of an ASAG system. Other approaches extracted the raw number of overlapping words and calculated several string similarity scores, e.g. cosine similarity and Lesk similarity~\cite{Dzikovska2012TowardsET}. In addition, the authors computed sub-tree matching based on lexical  features, in which they counted overlapping words and word stems. In~\cite{Meurers2011IntegratingPA}, the authors used features that are based on the overlapping words between the reference answer and the student answer.

Increasing the complexity and support of the overlap detection methods further improved performance. In~\cite{Jimnez2013SOFTCARDINALITYHT}, the authors proposed a method that computes sentence overlap in a hierarchical fashion, called SoftCardinality. It computes several similarity scores by considering different character q-grams  between words and sentences, and achieved state-of-the-art results on the SemEval data set. The achieved results demonstrated that a more fine-grained feature extraction technique  describes more complex characteristics of the text and, subsequently, improves the overall performance of the scoring system. Similar conclusions were drawn in~\cite{Heilman2013ETSDA}, where the authors proposed a method focused on the textual lexical similarity, computed by using character and word n-grams. In particular, the similarity was calculated taking into account the number of overlaps in the two sentences combined with string edit-distance features.
This approach uses additional features from~\cite{Heilman2012ETSDE}, which were computed according to the edit-effort it takes to align a student answer and a reference answer. The final score was predicted by a logistic regression classifier. In addition, domain adaption was used, where different features have individual weights for each domain.

\begin{landscape}
\begin{table}[!ht]
\Small
\renewcommand{\arraystretch}{1.1}
\caption{Details of the methods for ASAG based on lexical, syntactic and semantic features used in combinations with machine learning classifiers.}
  \begin{tabular}{ C{0.6cm}C{0.8cm}C{2.5cm}C{4.5cm}C{4.5cm}C{4cm}}
    \toprule
    ~ & ~ & ~ & \multicolumn{3}{c}{\bfseries Features} 
    \\ \cline{4-6} 
    \bfseries Ref. & \bfseries Year & \bfseries Classifier & \bfseries Lexical & \bfseries Syntactic  & \bfseries Semantic
    \\
    \midrule
    \cite{Mohler2011LearningTG} & 2011  & SVM & word overlap & dependency parsers, POS tags, n-grams with POS tags & WordNet, LSA 
    \\
    \midrule
     \cite{Meurers2011IntegratingPA} & 2011  & $k$NN & Bg-of-words, n-grams, word overlap & dependency parsers, POS tags, n-grams with POS tags, word synonyms & -
     \\
    \midrule
     \cite{Dzikovska2012TowardsET} & 2012 & Decision Tree & Bag-of-words, n-grams, word overlap & dependency parsers, POS tags, n-grams with POS tags & -
     \\
    \midrule
    \cite{Kouylekov2013CeliEA} & 2013 & SVM, Na\"{i}ve Bayes & word overlap & - & corpus-based statistical interrelations between words
    \\
    \midrule
    \cite{Levy2013UKPBIUSA} & 2013 & Na\"{i}ve Bayes & Bag-of-words, n-grams  & dependency parsers, POS tags, n-grams with POS tags & WordNet, LSA, ESA 
    \\
    \midrule
    \cite{Heilman2013ETSDA} & 2013 & Logistic Regression  & word/character overlap & - & -
    \\
    \midrule
    \cite{Ott2013CoMeTID} & 2013 & SVM, Logistic Regression & Bag-of-words, n-grams, word/character overlap & dependency parsers, POS tags, n-grams with POS tags, minimum edit distance & WordNet, LSA, ESA
\\
    \midrule
      \cite{Jimnez2013SOFTCARDINALITYHT} & 2013 & Bagged Decision Tree & word/character overlap  & -  & -
      \\
    \midrule
      \cite{Ramachandran2015IdentifyingPF} & 2015 & 
      Random Forest Regressor & Bag-of-words, n-grams, word overlap & dependency parsers, POS tags, word-graphs, phrase patterns & WordNet  
      \\
    \midrule
      \cite{Magooda2016} & 2016 & SVM & Bag-of-words, word/character overlap & - & word2vec, GloVE \\
    \midrule
      \cite{Sultan2016FastAE} & 2016 & Random Forest & bag-of-words, n-grams, word/character overlap & - & word2vec, GloVE 
      \\
    \midrule
       \cite{Roy2016AnIT} & 2016 & Logistic Regression & word overlap & - & word2vec, WordNet, LSA 
       \\
    \midrule
      \cite{Galhardi2018ExploringDF} & 2018 & Random Forests, Extreme Gradient Boosting & bag-of-words, n-grams, word/character overlap & dependency parsers, POS tags & WordNet (synonym similarity)
      \\
    \midrule
       \cite{Kumar2019GetIS} & 2019 & Random Forest & Bag-of-words, n-grams, word overlap 
       & dependency parsers, POS tags, n-grams with POS tags & word2vec, doc2vec
       \\
    \midrule
    \bottomrule
  \end{tabular}
  \label{tab:CAT1.features}
  \end{table}
\end{landscape}

\begin{landscape}
\begin{table}[!ht]
  \caption{Overview of performance of the ASAG methods based on hand-engineered features listed in Table~\ref{tab:CAT1.features}. We report accuracy unless otherwise specified. $\mathbf{\hat{F}}$ is the F1 score, $\mathbf{F_M}$ and $\mathbf{F_m}$ are the macro-averaged and micro-averaged F1 score, QW-K is the quadratic weighted kappa measure, RMSE is the root mean square error, and $\mathbf{\rho}$ is the Pearson's correlation coefficient. Performance measures are reported in the same precision as in the original papers.}
  \label{tab:CAT1.results}
\Small
\renewcommand{\arraystretch}{1.1}
  \begin{tabular}{@{\extracolsep{4pt}} C{0.6cm}C{0.8cm}C{2.5cm}C{1.2cm}C{1.2cm}C{1.2cm}C{1.2cm}C{1.2cm}C{1.2cm}C{2cm}C{2cm}   }
    \toprule
    ~ & ~ & ~ & \multicolumn{3}{c}{\bfseries SciEntsBank} & \multicolumn{3}{c}{\bfseries Beetle} & \bfseries Texas2011 & \bfseries Other 
    \\ \cline{4-6} \cline{7-9}
    \bfseries Ref. & \bfseries Year & \bfseries Classifier & \bfseries 2-way & \bfseries 3-way & \bfseries 5-way & \bfseries 2-way & \bfseries 3-way & \bfseries 5-way & ~ & ~ \\
    \midrule
    \cite{Mohler2011LearningTG} & 2011 & SVM & - & - & - & - & - & - & $0.518$~($\mathbf{\rho}$) $0.998$~(RMSE) & -\\
    \midrule
     \cite{Meurers2011IntegratingPA} & 2011  & $k$NN & - & - & - & - & - & - & - & $0.79$ English~Dev.~Corpus\\
    \midrule
     \cite{Dzikovska2012TowardsET} & 2012  & Decision Tree & - & - & $0.29$~($F_M$) $0.42$~($F_m$) & - & - & - & - & -\\
    \midrule
    \cite{Kouylekov2013CeliEA} & 2013 & SVM, Na\"{i}ve Bayes & $0.612$ & $0.55$ & $0.421$ & $0.648$ & $0.523$ & $0.464$ & - & - \\
    \midrule
    \cite{Levy2013UKPBIUSA} & 2013 & Na\"{i}ve Bayes & $0.696$ ($\hat{F}$) & $0.606$ ($\hat{F}$) & $0.464$ ($\hat{F}$) & - & - & - & - & - \\
    \midrule
    \cite{Heilman2013ETSDA} & 2013 & Logistic Regression & - & - & $0.524$ ($\hat{F}$) & - & - & $0.659$ ($\hat{F}$) & - & -\\
    \midrule
    \cite{Ott2013CoMeTID} & 2013 & SVM, Logistic Regression & $0.684$ & $0.612$ & $0.486$ & $0.77$ & $0.624$ & $0.588$ & - & - \\
    \midrule
      \cite{Jimnez2013SOFTCARDINALITYHT} & 2013 & Bagged Decision Tree & $0.726$ & $0.649$ & $0.527$ & $0.724$ & $0.538$ & $0.513$ & - & - \\
    \midrule
      \cite{Ramachandran2015IdentifyingPF} & 2015 & Random Forest Regressors & - & - & - & - & - & - & $0.61$~($\mathbf{\rho}$) $0.86$~(RMSE) & $0.78$~(QW-K) ASAP-SAS \\
    \midrule
      \cite{Magooda2016} & 2016 & SVM & $0.605$ ($\hat{F}$) & - & $0.48$ ($\hat{F}$) & - & - & - & - & -\\
    \midrule
      \cite{Sultan2016FastAE} & 2016 & Random Forest & - & - & $0.56$ ($\hat{F}$) & - & - & - & $0.85$~($\mathbf{\rho}$) $0.63$~(RMSE) & -\\
    \midrule
       \cite{Roy2016AnIT} & 2016 & Logistic Regression & - & - & $0.565$ ($\hat{F}$) & - & - & - & $0.82$~(RMSE) & -\\
    \midrule
      \cite{Galhardi2018ExploringDF} & 2018 & Random Forest, Extreme Gradient Boosting & $0.775$ & $0.719$ & $0.59$ & $0.81$ & $0.643$ & $0.644$ & - & - \\
    \midrule
       \cite{Kumar2019GetIS} & 2019 & Random Forest & - & - & - & - & - & - & - & $0.791$~(QW-K) ASAP-SAS\\
    \bottomrule
  \end{tabular}
\end{table}
\end{landscape}

\subsection{Syntactic features}

Syntactic features are able to quantify important information about the meaning of a sentence, as they detect and describe the roles and dependencies of the words in it. The ability to characterize the relation between words enables to perform inference about the meaning of a textual answer~\cite{Lee2017ExploringLA}.
Basic approaches to extract syntactic features from text are using parse trees and part-of-speech tagging (POS tags) or dependency n-grams~\cite{Popovic2011MorphemesAP}. Dependency n-grams are derived from the syntactical relation between words by grouping, for instance, verb and subject together. In~\cite{Levy2013UKPBIUSA}, syntactic features were computed by generating n-grams that consist of combinations of POS tags. The dependency of words in sequences was subsequently used to assess the content of an answer and evaluate its similarity with a reference answer. This method worked based on the concept of paraphrasing a sentence, the result of which consists of a sentence having different wording but similar meaning. 
In~\cite{Dzikovska2012TowardsET}, the authors used different similarity scores based on POS tags. Other approaches focused on the syntactical dependency, used parse trees, and extracted graph alignments features~\cite{Mohler2011LearningTG}. Similarity measures were enriched by generating word pairs with similar POS tags and calculating the corresponding similarity. The underlying assumption was that similarly structured answers are more likely to have a similar meaning.

\subsection{Semantic features}
Lexical features can not capture the semantic content of sentences, and syntactic features are able to catch such characteristics only to a limited extent. Thus, more sophisticated features were designed exploiting knowledge-bases to recognize the meaning of words therefore more robustly computing similarity among sentences. These methods used knowledge sources like WordNet~\cite{Miller1995WordNet} and computational approaches based on Latent Semantic Analysis (LSA)~\cite{Landauer1998LSA} and Explicit Semantic Analysis (ESA)~\cite{Gabrilovich2007ESA}, combined with different similarity metrics.
WordNet models the semantic relation between words inducing synonyms and hyponyms. Many methods used Wordnet in combination with similarity metrics to better incorporate the semantic meaning of words. 
LSA is a corpus-based similarity method, in which words are represented as vectors in a multi-dimensional semantic space~\cite{Kaur2005ACO}. This method gained popularity and was demonstrated to perform better than word and n-gram vectors~\cite{Mohler2009}. LSA was used to estimate the similarity between words and combined with a word-weighting factor to enhance the relevance of specific words in~\cite{Kouylekov2013CeliEA}. ESA was designed to use the knowledge extracted from Wikipedia~\cite{Gabrilovich2007ComputingSR} and was demonstrated to perform comparably to LSA or to outperform it in some cases. 
In~\cite{Levy2013UKPBIUSA}, the authors included the semantic similarity by using predefined context vectors based on WordNet. They proposed to first parse the answers according to the dependencies of words within the sentences, and subsequently calculate the semantic similarity using the closest common ancestor and shortest path length using ESA and WordNet.
In~\cite{Ott2013CoMeTID}, WordNet was also used to extract similarity information between the reference and student answer. The authors highlighted the importance of using different features to extract semantic information and weight them for an effective classification task. 


\section{Deep learning methods}
\label{sec:DL}
The deep learning developments in the ASAG field align with the methodological advances in the field of NLP. We organize the deep learning methods for ASAG in three categories, that correspond to the historical development of NLP and the relation to the text representation methods. 

The first category contains methods based on word embedding models (e.g. Word2Vec~\cite{Mikolov2013EfficientEO}). These models compute representations that transform similar words into vectors that are close to each other in an embedded latent space, and generate sentence embeddings by either summing or averaging the single word embeddings. The word  and sentence embeddings are able to capture semantic information in textual data more effectively than previous hand-engineered features~\cite{Hightower}.
Methods in the second category deploy recurrent neural networks (RNNs), of which those based on long short-term memory (LSTM) networks are very popular, to model the sequential characteristics of textual data. 
These methods are able to capture semantic properties of the text by considering word sentences of different lengths, and longer-range relationships between words in sentences. This allowed the prediction models to carry out more robust and effective inferences on given answers~\cite{Saha2019JointML}.
The third category consists of methods that deploy attention-based mechanisms, able to describe long-range relationships between words in a sentence. Different from RNNs, attention-based methods do not need to explicitly model the sequential characteristics of words in sentences, but are able to process longer sentences in parallel. These architectural advancements were determined by the use of several self-attention components, each of which captures a specific relation between words. Architectures based only on attention are called Transformers~\cite{Vaswani2017AttentionIA}. 

We elaborate on the details of the three categories of methods in the remainder of this section, addressing the benefits, drawbacks, and performance of the learned text representations. In Table~\ref{tab:DLdesc}, we provide an overview of the methodological aspects of the reviewed papers, while in Table~\ref{tab:DLresults} we summarize their performance results on benchmark data sets.

\subsection{Word embeddings}
Deep learning methods for ASAG based on word embeddings  encode semantically similar words to close points in a latent space~\cite{Roy2016AnIT,Magooda2016,Sultan2016FastAE}. The general success of word embeddings is attributable to their ability to describe rich semantic features of text. These methods were demonstrated successful to improve the evaluation of word similarity, but did not clearly outperform previous methods for representation of entire sentences in ASAG systems. In~\cite{Magooda2016}, for instance, the authors compared different similarity measures on word vector representations of text obtained using pre-trained embedding models, such as Word2Vec~\cite{Mikolov2013EfficientEO} and GloVe~\cite{Pennington2014GloveGV}. They found that word and sentence embeddings achieved below-average results on the SemEval 5-way task. Other approaches that solely focused on embeddings did not outperform the models with hand-engineered features on ASAG tasks~\cite{Gomaa2019Ans2vecAS,Magooda2016}. 

To compensate for the shortcomings of word embeddings techniques to represent entire sentences, several methods that combine word embeddings with hand-engineered features were designed. The method proposed in~\cite{Roy2016AnIT}, for instance, combined an external knowledge-like paraphrase database and WordNet with syntactical similarity features. It achieved top performance results on the SciEntsBank dataset, with improved generalization capabilities. Other approaches, such as that proposed in~\cite{Sultan2016FastAE}, computed whole-sentence representations by combining (sum or average) single word embeddings. 
Lexically similar word pairs were obtained using an external paraphrase data set, with labeled graded short answers. The semantic similarity of words was computed using pre-trained word embeddings  and the cosine similarity function, while the representation of entire sentences was computed by summing up the word embedding vectors. This approach resulted in relatively good results on a proprietary data set, but highlighted the need for further developments of sentence representations.

\begin{landscape}
\Small
  \begin{longtable}[H]{C{0.8cm}C{0.8cm}C{0.8cm}C{3cm}C{5cm}C{5cm}}
  \caption{Overview of deep learning methods for ASAG, the type of semantics that they learn from the text, and the focus of their architecture. \label{tab:DLdesc}} \\
    \toprule
    \bfseries Ref. & \bfseries Cat. & \bfseries Year & \bfseries Model & \bfseries Features & \bfseries Focus\\
    \midrule
   \cite{Riordan2017InvestigatingNA}  &  DL3  & 2017 & LSTM+CNN with attention & Combination of sequence representation from different model architecture (CNN, LSTM) & Complex and stacked model to leverage benefit from different models \\
    \midrule
    \cite{Kumar2017EarthMD}  & DL2  & 2017 & BiLSTM & Sequence representation with BiLSTM & Complex model architecture and included data augmentation \\
    \midrule
    \cite{Saha2018SentenceLO}  & DL1  & 2018 & Random Forest & Sentence representation combined with engineered features & Combination of sequence representation and engineered features with focus on domain adaptation \\
    \midrule
    \cite{Kumar2019GetIS}  & DL1  & 2019 & Random Forest & Sentence and word representation & Combination of sequence representation and various engineered features \\
    \midrule
    \cite{Wang2019InjectRI}  & DL3  & 2019 & BiLSTM & Attention based sentence representation & Complex structure and incorporation of engineered features \\
    \midrule
    \cite{Saha2019JointML}  & DL2  & 2019 & BiLSTM & Combination of different sentence representations (domain-specific and unspecific) & Complex structure and domain adaptation \\
    \midrule
    \cite{Gomaa2019Ans2vecAS}  & DL1  & 2019 & Logistic Regression & Sequence representation of words and sentences & Simple application of transfer learning \\
    \midrule
    \cite{Sung2019ImprovingSA}  & DL3  & 2019 & BERT & Attention based sequence representation & Simple application of transfer learning \\
    \midrule
    \cite{Qi2019AttentionBasedHM}  & DL3  & 2019 & BiLSTM+CNN & attention based on BiLSTMs and CNNs & Complex and stacked BiLSTM and CNN \\
    \midrule
    \cite{Liu2019AutomaticSA}  & DL3  & 2019 & Multiway-attention transformer & attention & Complex structure with incorporation of different attention mechanism \\
    \midrule
    \cite{Sung2019PreTrainingBO}  & DL3  & 2019 & BERT & attention & Transfer learning with domain adaptation \\
    \midrule
    \cite{Zhang2019AnAS}  & DL2  & 2019 & LSMT & CBOW word vectors & Incorporation of domain-specific knowledge and domain-general knowledge from wikipedia \\
    \midrule
    \cite{Zhang2020GoingDA}  & DL2  & 2020 & DBN & Application of Gaussian mixture model (GMM) and Cartesian product for feature composition & Complex feature engineering by incorporating different feature extraction methods \\
    \midrule
    \cite{Camus2020InvestigatingTF}  & DL3  & 2020 & RoBERTa & Attention based sequence representation & Targeted transfer learning and incorporation of different learning methods (cross-lingual, NLI specific model training) \\
    \midrule
    \cite{Condor2020ExploringAS}  & DL3  & 2020 & BERT & Attention based sequence representation & Simple application of transfer learning \\
    \midrule
    \cite{Sahu2020FeatureEA}  & DL1  & 2020 & stacked regression ensemble & Compilation of various engineered features & Ensemble of eight different regression models via multi-layer perceptron and various engineered features \\
    \midrule
    \cite{Gaddipati2020ComparativeEO}  & DL2  & 2020 & Stacked BiLSTM (ELMo) & Bidirectional LSTM based sequence representation & Simple application of transfer learning from pretrained model \\
    \bottomrule
  \end{longtable}

\end{landscape}

 \begin{landscape}
     \Small
     \renewcommand{\arraystretch}{1.3}
   \begin{longtable}{@{\extracolsep{2pt}} C{0.6cm}C{0.6cm}C{0.8cm} C{1.4cm}C{1.8cm}C{1.4cm} C{1.4cm}C{1.2cm}C{1.4cm} C{2.2cm} C{2.2cm}   }
  \caption{
  Summary of the performance results of the ASAG methods based on deep learning listed in Table~\ref{tab:DLdesc}. We report accuracy unless otherwise specified. $\mathbf{\hat{F}}$ is the F1 score, QW-K is the quadratic weighted kappa measure, C-K is the Cohen's Kappa, RMSE is the root mean square error, and $\mathbf{\rho}$ is the Pearson's correlation coefficient. Performance measures are reported in the same precision as in the original papers.
   \label{tab:DLresults}}\\
     \toprule
     \multicolumn{3}{c}{~} & \multicolumn{3}{c}{\bfseries SciEntsBank} & \multicolumn{3}{c}{\bfseries Beetle} & \bfseries Texas2011 & \bfseries Other      \\ \cline{4-6} \cline{7-9}
     \\
    \bfseries Ref. & \bfseries Cat. & \bfseries Year & \bfseries 2-way & \bfseries 3-way & \bfseries 5-way & \bfseries 2-way & \bfseries 3-way & \bfseries 5-way & ~ & ~\\
    \midrule
    \cite{Riordan2017InvestigatingNA}  & DL3  &  2017 & $0.712$ & - & $0.533$ & $0.79$ & - & $0.633$ & $0.518$~($\mathbf{\rho}$) $0.998$~(RMSE) &  $0.723$~(QW-K) ASAP-SAS\\
    \midrule
    \cite{Kumar2017EarthMD}  & DL2  &  2017 & - & $0.634$~(MAE) $0.904$~(RMSE) $0.316$~($\mathbf{\rho}$) & - & - & - & - & $0.61$~($\mathbf{\rho}$), $0.77$~(RMSE) & -\\
    \midrule
    \cite{Saha2018SentenceLO}  & DL1  &  2018 & $0.752$ & $0.654$ & $0.540$ & - & - & - & $0.57$~($\mathbf{\rho}$) $0.902$~(RMSE) & ~ \\
    \midrule
     \cite{Kumar2019GetIS}  & DL1  & 2019 & - & - & - & - & - & - & - & $0.791$~(QW-K) ASAP-SAS\\
    \midrule
    \cite{Wang2019InjectRI}  & DL3  &  2019 & - & - & - & - & - & - & - & $0.77$~(QW-K) ASAP-SAS \\
    \midrule
    \cite{Saha2019JointML}  & DL2 & 2019 & $0.803$~($\hat{F}$) & $0.744$~($\hat{F}$) & $0.656$~($\hat{F}$) & - & - & - & - & $0.721$~($\hat{F}$) Large Scale Industry Dataset\\
    \midrule
    \cite{Gomaa2019Ans2vecAS}  & DL1  & 2019 & - & - & $0.503$~($\hat{F}$) & - & - & - & $0.63$~($\mathbf{\rho}$) $0.91$~(RMSE) & -\\
    \midrule
    \cite{Sung2019ImprovingSA}  & DL3  & 2019 & - & $0.68$~($\hat{F}$) & - & - & - & - & - & -\\
    \midrule
    \cite{Qi2019AttentionBasedHM}  & DL3  &  2019 & - & - & - & - & - & - & - & $0.969$~(Acc) $0.969$~($\hat{F}$) Chinese~data \\
    \midrule
    \cite{Liu2019AutomaticSA}  & DL3  & 2019 & - & - & - & - & - & - & - & $0.889$~(Acc) $0.944$~(AUC) Real~world~K-12\\
    \midrule
    \cite{Sung2019PreTrainingBO}  & DL3  & 2019 & - & - & - & - & - & - & - & $0.799$~(Acc) Large scale industry dataset\\
    \midrule
    \cite{Zhang2019AnAS}  & DL2  & 2019 & - & - & - & - & - & - & - & $0.626$~(QW-K) ASAP-SAS\\
    \midrule
    \cite{Zhang2020GoingDA}  & DL2  & 2020 & - & - & - & - & - & - & - & $0.850$~(Acc) $0.830$~($\hat{F}$) \hbox{Cordillera}\\
    \midrule
    \cite{Camus2020InvestigatingTF}  & DL3  & 2020 & - & 0.718 & - & - & - & - & - & -\\
    \midrule
    \cite{Condor2020ExploringAS}  & DL3  & 2020 & - & - & - & - & - & - & - & $0.76$~(Acc)  $0.684$~(C-K) DT-Grade\\
    \midrule
    \cite{Sahu2020FeatureEA}  & DL1  & 2020 & - & - & 0.746 & - & - & $0.666$ & - & -\\
    \midrule
    \cite{Gaddipati2020ComparativeEO}  & DL2  & 2020 & - & - & - & - & - & - & $0.485$~($\mathbf{\rho}$) $0.978$~(RMSE) & -\\
    \bottomrule
  \end{longtable}
 \end{landscape}


Previous works showed that ASAG systems perform well when word-embeddings and hand-engineered features are combined, while the sole use of word-embeddings does not contribute to consistently good results~\cite{Gomaa2019Ans2vecAS,Magooda2016}. For instance, in~\cite{Kumar2019GetIS}, good performance results were achieved by constructing a large feature set by combining hand-engineered and learned features. The authors proposed a method that combines commonly used text representations, such as Word2Vec, Doc2Vec, POS tagging, n-gram overlaps, with features that capture the diversity and style of formulation of the student answers. The design of this method was based on the hypothesis that the use of a more sophisticated textual form, in terms of the language used, is an indication for an answer of higher quality.
They achieved state-of-the-art results on the ASAP-SAS data set and demonstrated that the Word2Vec and Doc2Vec embeddings, combined with features for quantification of text overlaps and weighted keywords play a relevant role in the automatic grading of answers. By including features that represent the complexity of the answers, the accuracy of the proposed system improved significantly. On the one hand, researchers agree on the importance and descriptive capabilities of embeddings~\cite{Kumar2019GetIS}. On the other hand, it was demonstrated that well hand-engineered features that capture complex properties of the text based on prior knowledge can be beneficial for the overall system performance: knowledge extracted from the text by embedding-based representations is complementary to that of previously proposed syntactical, semantic, and lexical features~\cite{Roy2016AnIT}.

\subsection{Sequence-based models}
Sequential machine learning models were applied to ASAG in order to improve  the quality of the learned features and robustness of the word and sentence representation. 
Kumar et al.~\cite{Kumar2017EarthMD} proposed to use a siamese bidirectional-LSTM architecture combined with a pooling layer that uses earth-mover distance. The authors compared the pairwise distance between the latent vectors of the reference answer and the student answer. They showed that the extraction of semantic textual features using sequence-based models improves the quality of the learned representation and the performance of the automated grading systems.

Further focus was given to the training of sequential models for NLP and their adaptation to ASAG problems. Specifically, fine-tuning of pre-trained models for the analysis of text sequence from other, more general domains to the task of ASAG was explored using techniques for transfer learning~\cite{Cai2019,Tsiakmaki2020}. Transfer learning provides the possibility of exploiting features that are learned from large corpora of text data, and that have more powerful semantic representation capabilities. In this context, the Universal Sentence Representation model was often used in transfer learning settings to extract word representations~\cite{conneau-etal-2017-supervised}. The authors trained a bi-directional LSTM network on the large Stanford Natural Language Inference corpus~\cite{conneau-etal-2017-supervised} and subsequently adapted the feature extraction on the SemEval ASAG tasks~\cite{Saha2018SentenceLO}. The authors achieved state-of-the-art performance on the SemEval tasks (see Table~\ref{tab:DLresults}), by computing the semantic similarity between word- and sentence-embeddings of the reference and student answers computed via the adapted models. In~\cite{Saha2018SentenceLO}, hand-engineered features were also used for different question types. These results indicate  that combining learned sequential-based and hand-engineered features increase the performance of ASAG methods. 

\begin{figure}[!t]
    \centering
    \includegraphics[width=0.9\columnwidth]{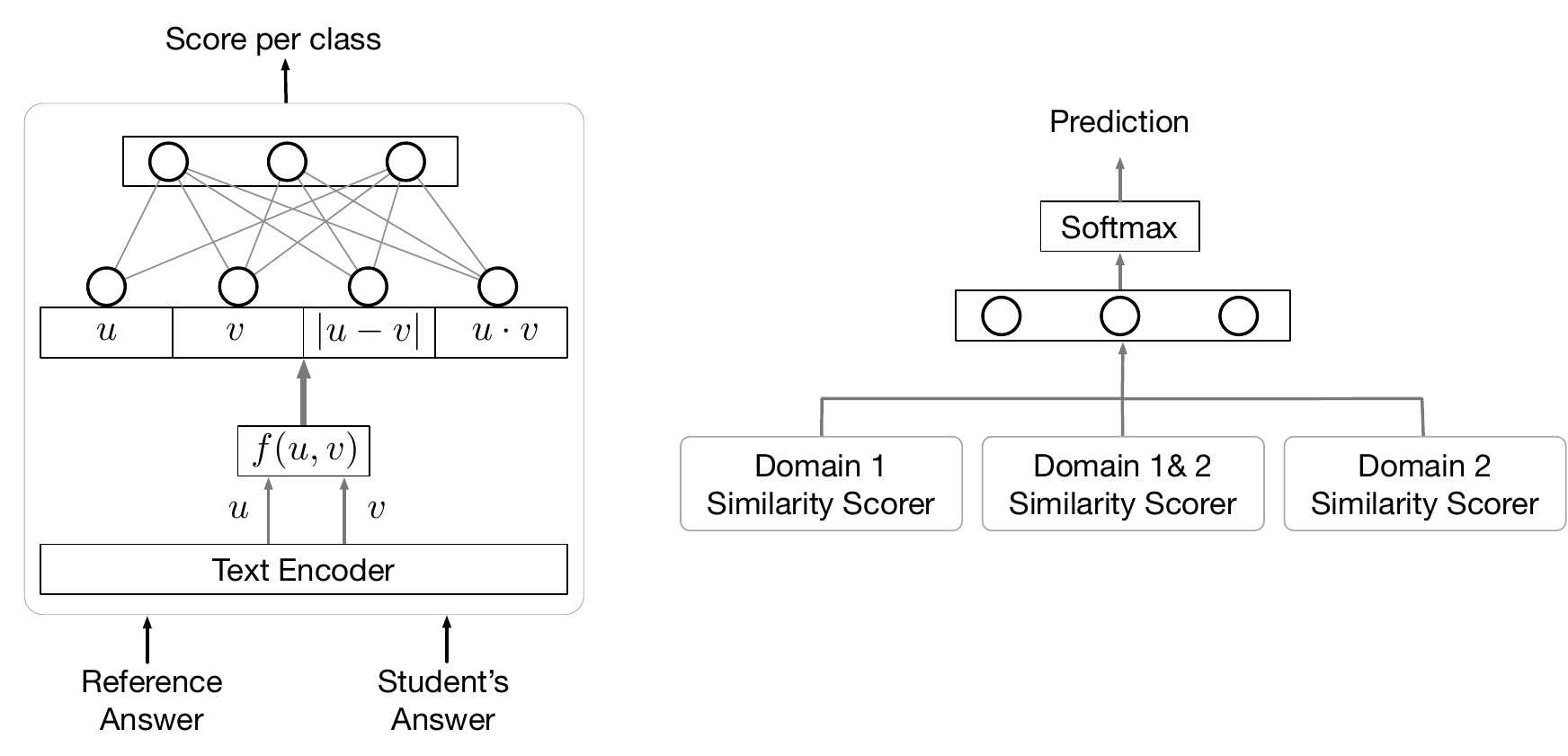}
    \caption{Architecture of the method published in~\cite{Saha2019JointML}. The authors used domain adaptation by training different similarity scorers on domain-specific subsets of answers and on a domain-independent data set.}
    \label{fig:model_jointml}
\end{figure}

Domain adaptation of pre-trained sequence-based models was also explored in~\cite{Zhang2019AnAS}. The authors trained an LSTM with focus on the incorporation of domain-specific knowledge and domain-general knowledge from Wikipedia. In~\cite{Saha2019JointML}, a combination of domain adaptation and transfer learning techniques were used (see Figure~\ref{fig:model_jointml}). The authors argued that the performance of systems that rely on textual similarity, paraphrasing, or entailment depends on the domain in which they are applied. 
They, thus, propose a deep learning architecture that deploys an encoder with a BiLSTM layer to embed the reference answer and the student answer into a vector representation, trained on questions and answers from multiple domains. To compute answer similarity, the authors propose to train a generic scoring model on answers from all considered domains, and specific scoring models trained on answers drawn from specific domains. The final prediction on the correctness of a test answer is computed by summing up the scores given by the generic and domain-specific scorers. This method achieved state-of-the-art results on the SemEval tasks, demonstrating that a combined training of a prediction model  with multi-domain and domain-specific data contributes to an increase in the overall performance of an ASAG system. This indicates that the performance results of models for ASAG tasks are influenced by the capabilities of the underlying word-embedding models to effectively capture the semantic properties of words and sentences. Thus, the use of models that can integrate longer sequential relations in textual input proved to enrich the semantic information of the embeddings, and improved model performance. However, this is usually accompanied by an increase in the complexity of the models ~\cite{Sahu2020FeatureEA,Zhang2020GoingDA}.

Ensembles of sequence-based models were also considered as a way to enhance the representation power of learned features~\cite{Zhang2020GoingDA,Sahu2020FeatureEA}. Both works used an ensemble of eight different stacked regression models whose predictions were combined using a multi-layer perceptron. The authors included summaries of the responses in the training data, which contributed to learning more robust models. 

\subsection{Attention-based models} 
More recent methods for ASAG explored more sophisticated feature representations, computed with attention-based and transformer models, to capture better and more descriptive semantic and structural characteristics from text. Attention enables the calculation of the relation and relative importance between each word within a sentence. The attention mechanism allows to model the dependencies of words and their importance for the prediction task at a longer range in sentences. The modeling does not take into account the sequentiality of words explicitly. 

Architectures relying purely on attention mechanisms were introduced in~\cite{Vaswani2017AttentionIA}, and are called transformers. They consist of an encoder-decoder structure able to characterize long-range characteristics and dependencies in sequential data. In transformer architectures, the attention mechanism is modeled via multiple parallel attention-head components, each of them able to learn different dependencies. 

Attention-based models differ in their general architecture. In~\cite{Riordan2017InvestigatingNA}, LSTMs and Convolutional Neural Networks (CNNs) were combined with attention mechanisms, and they were demonstrated to outperform previous methods. In particular, bidirectional LSTMs augmented with attention  obtained competitive results for ASAG problems (see Table~\ref{tab:DLresults}). According to the authors, the choice of the input embeddings and 
the correct fine-tuning of the pre-trained models on the task at hand is crucial to obtain good results. Several further works focused on consistently applying transfer learning techniques, with the aim of fine-tuning the feature embedding space towards the domain of interest. For this purpose, more complex architectures, made of stacked or ensemble networks and attention modules were proposed~\cite{Camus2020InvestigatingTF,Pribadi2017AutomaticSA,Riordan2017InvestigatingNA}.

Other methods were based on the use of transformer models~\cite{Vaswani2017AttentionIA}. Their success and high-performance results are attributable to the high parallelization of the computations, which allows to train models on larger data sets, and the ability of modeling long-range dependencies. Further research work thus focused on exploring the use of transformers to compute text embeddings, and their effects on NLP and ASAG tasks ~\cite{Camus2020InvestigatingTF}.
In~\cite{Sung2019ImprovingSA}, the Bidirectional Encoder Representation from Transformers (BERT) model was fine-tuned on an ASAG task, achieving state-of-the-art performance. The authors observed that a BERT model can be fine-tuned using only a few labeled examples on a target task and achieve very good performance, although it suffers from a lack of cross-domain generalization. 

The sparse characteristics of data in benchmark data sets required the use of transfer learning and domain adaptation techniques also for fine-tuning transformer models and tailoring the learning process to the task at hand or its specific domain. In~\cite{Sung2019PreTrainingBO}, the effect of different training approaches was investigated for ASAG. The authors showed that utilizing unstructured domain text data and question-answer pairs to fine-tune the models resulted in better results than using task-specific data only. The method proposed in~\cite{Sung2019PreTrainingBO} achieved higher results than all previous works on the SemEval 3-way task, although results on the SemEval 5-way task were not reported. 
In~\cite{Camus2020InvestigatingTF}, the authors compared the performance of several state-of-the-art transformers fine-tuned using different learning strategies, such as cross-lingual and task-specific training, and noticed a consistent enhancement of the results on ASAG tasks. 

\begin{figure}[!t]
    \centering
    \input{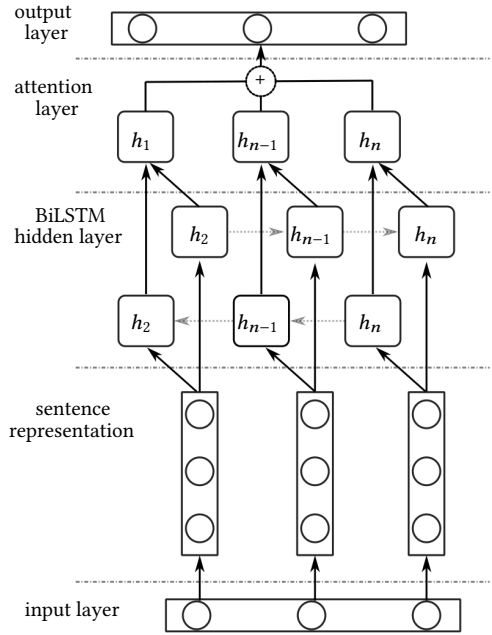}
    \caption{Sketch of the model architecture proposed in~\cite{Qi2019AttentionBasedHM}. The authors proposed a combined use of a CNN and a BiLSTM, together with an attention layer, which provides a refinement of the prediction before the output layer is computed.}
    \label{fig:blstm-cnn}
\end{figure}

\begin{figure}[!t]
    \centering
    \includegraphics[width=0.8\columnwidth]{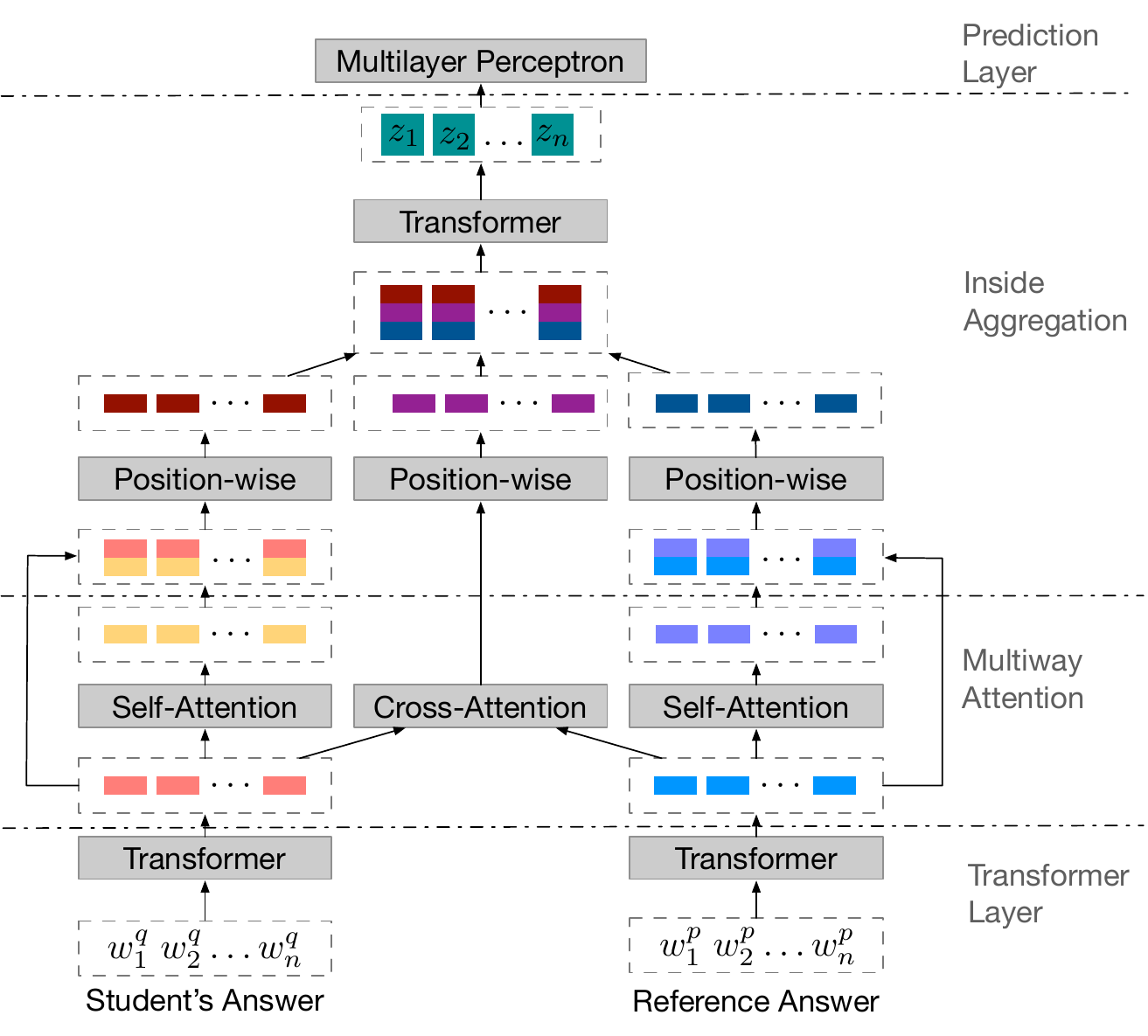}
    \caption{Sketch of the model architecture proposed in~\cite{Liu2019AutomaticSA}. The authors combined different representations of the student and reference answers. Subsequently, these representations are aggregated at the prediction stage.}
    \label{fig:model-architecture-attention}
\end{figure}

Since attention is instrumental for the success of transformer models~\cite{conneau-etal-2017-supervised,Peters2018DeepCW,Radford2018ImprovingLU,Howard2018UniversalLM,Devlin2019BERTPO,Yang2019XLNetGA} and their application to ASAG, researchers focused on how it can be used optimally in a model. In~\cite{Qi2019AttentionBasedHM}, for instance, an attention layer was combined with a BiLSTM and a CNN architecture to compute question-answer representations. The structure of the model architecture is illustrated in Figure~\ref{fig:blstm-cnn}. Such embeddings were concatenated and used for scoring the student answers. With this approach, the authors leveraged textual semantic features and long-term dependencies in text and obtained good performance results. 
Furthermore, in~\cite{Wang2019InjectRI}, it was shown that attention can be used to extract key elements from the student answers. The authors demonstrated that the performance could be improved by considering a larger observation window on the text rather than using only word-level attention. Combinations of different attention mechanisms were explored in~\cite{Liu2019AutomaticSA}. As illustrated in Figure~\ref{fig:model-architecture-attention}, the model calculated separate representations for the student answer, the reference answer, and their cross-attention. These different representations are then aggregated position-wise and used as input representations for a basic transformer model. Their approach achieved overall good results on the Real World K-12 data set (see Table~\ref{tab:DLresults}).

The use of attention-based models contributed to a substantial increase in performance on ASAG tasks, which is attributable to the more powerful representations learned using these models. The analysis and comparison of longer text data, however, remains a challenging aspect of ASAG systems. We discuss the challenges and outlook of further developments in this field in the following section.

\section{Discussion}
\label{sec:discussion}

The field of ASAG has witnessed a transition from methodologies based on careful design of hand-engineered text features towards feature learning architectures based on deep learning. This evolution has been strongly influenced by the progress made in the field of NLP. Transfer learning from very big models trained on generic NLP tasks to target tasks has become a common practice. This, however, has benefits but also creates challenges in the field of ASAG. 

Word embeddings, sequential models, and attention-based and transformers models have progressively contributed to improving the semantic richness and descriptive power of textual representation. This resulted in impressive progress in the field of NLP~\cite{Sahu2020FeatureEA}, mainly due to the ability to exploit the relations and the diversity of structures of vast amounts of data. 
The performance of ASAG methods has, however, not witnessed the same level of improvement. This is mainly attributable to the fact that the ASAG benchmark datasets are sparse, and very complex models developed for general NLP tasks do not generalize well to this application field. Although transfer-learning approaches are shown to be promising~\cite{Sung2019ImprovingSA}, fundamental aspects of ASAG like sparsity and domain differences need to be addressed explicitly. 

Transformers and attention-based models have achieved astonishing results in NLP~\cite{Liu2019RoBERTaAR}. For ASAG tasks, however, a fine-tuned transformer model alone does not achieve the highest performance on benchmark datasets. We hypothesize that these complex embeddings models are not able to effectively disentangle the semantically rich information contained in short answers. The best-performing methods, indeed, combine complex embeddings representation computed by transformers with sets of hand-crafted features that are specifically designed to address particular aspects and characteristics of short answers. 
Currently, the research community of NLP and ASAG is investigating ways to optimize the embeddings of paragraphs of medium size which best represent the semantic information contained in the student answers.  
Particular attention is given to: 
\begin{inparaenum}[\itshape a\upshape)]
    \item combination of different features to exploit their ability to 
    \item using multi-classifier systems (e.g. stacked architectures, ensembles, and hybrid models) to exploit eventual complementary modeling and predictive capabilities, and
    \item optimizing the training process to influence what semantic information the model focuses more on (e.g. different domain adaptation strategies, general or task-focused pre-training).
\end{inparaenum}

\subsection{Open challenges}
We have recognized a number of challenges that ASAG methods have to deal with, particularly because of the peculiar characteristics of the task at hand. In the following paragraphs, we discuss them.

\paragraph{\bfseries Semantic understanding.} Existing methods have difficulties to model the rich semantic content of short answers effectively. This is due to the fact that, usually, short answers are written in such a way that a lot of information is provided in few sentences and in a very concise way. Therefore, ASAG systems  require methods for better and deeper contextual and semantic understanding of the text, for which the application of Natural Language Understanding (NLU) techniques~\cite{Navigli2018NLU} should be investigated.

\paragraph{\bfseries Linguistic variations.}
The analysis of student answers presents several challenges related to the way sentences are formulated. A concept or intention can be expressed in different ways, such as with different words or with the construction of sentences~\cite{Roy2016AnIT}.
Furthermore, answers may lack a complete sentential form or be deliberately written to trick the automatic evaluation systems. 
The grading of non-sentential answers is challenging because they might not be in conformance with the structural pattern of the reference answer or grammatically incorrect~\cite{Saha2018SentenceLO}. Grading systems need to be able to take into account these cases and interpret or infer complete answers from fragments.
In addition, most of the existing methods are not able to distinguish between nonsense and relevant answers. Due to the intended automated nature of the system, it is crucial to detect if a student is trying to trick the system by answering the question with consecutive incoherent keywords, instead of formulating a complete answer.

\paragraph{\bfseries Details of questions and reference answers.} Training ASAG models rely on the correctness and reliability of the reference answers and labels. Some reference answers used to train the models might be too brief, not containing enough details, or even not completely meaningful. This creates issues for automated grading algorithms, which have to be trained to be robust to data uncertainty. 
Likewise, ASAG models should also be able to handle different question types and answer expectations. For instance, some questions may require providing a definition whereas other questions expect a more complex and detailed answer~\cite{Roy2016AnIT}. Especially open-ended questions are challenging since they expect the students to express their own thoughts based on facts. Hence, a reference answer might not be suitable because correct answers may vary a lot~\cite{Zhang2019AnAS}.

\paragraph{\bfseries Generalization across domains and answers. }

ASAG systems are required to maintain consistent levels of performance when evaluating the answers to questions from different domains~\cite{Saha2018SentenceLO}. This is not a straightforward task to be evaluated, as existing data sets are biased towards certain domains, or reference answers with higher scores~\cite{Wang2018IdentifyingCI}. The sparsity of available benchmark data sets thus poses challenges to effectively training models with robust generalization abilities. Furthermore, for a given question, multiple reference answers can be correct, meaning that there is no absolute gold standard to compare the student answers with~\cite{Roy2016AnIT}. Training models for ASAG needs to take into account the sparse characteristics of the available data and possible noisy labels, in order to avoid overfitting.

\subsection{Outlook and future work}

Deep learning approaches have been beneficial for the improvement of the performance of the ASAG system and have been demonstrated to effectively complement the text representation capabilities of methods based on hand-engineered features. To further improve the possibilities and extent of application of deep learning methods for grading of short answers, necessary steps have to be taken that address the specific problems and challenges presented above. 

The available benchmark data sets are rather sparse and are not representative enough of the  variability of questions and short reference answers in different domains. This hinders the generalization capabilities of learning-based methods, which are thus subject to overfitting. This makes tasks like the extension of existing data sets, data augmentation~\cite{Wei2019}, and the creation of synthetic data via generative models~\cite{Tevet2019} very relevant to promote future progress in the field. 

The explainability~\cite{Guidotti19} and robustness~\cite{wang2021robust} of deep learning-based models are also important aspects that require further research. These can be indeed instrumental characteristics to assess whether a model is capable of robust decision making and can be applied successfully to grade answers in different topic domains. Furthermore, being able to consistently explain the decisions taken by a certain model also supports the analysis of what words or semantic constructs the score is determined by. Performance analysis and model inspection are thus complementary aspects that should be jointly taken into account for the evaluation of the developed methods, instead of only focusing on improving benchmark results. Promising research areas concern the identification of the points of strength of existing methods, to determine sets of methods that are suitable to be used for specific tasks and subsequently design and test ensemble methods for multi-domain cases. However, in order to achieve this goal, techniques to explain the model predictions should be put in place.
These aspects are strictly linked with the need of characterizing the robustness of the prediction made by the trained models to variations of the test answers~\cite{CATGEN}. 
A research objective that needs to be taken into account is to study how sensitive ASAG models are to changes in the answers of students, in order to avoid that they can be tricked by using certain keywords or by swapping the order of words. This has to be coupled with a more extensive use of NLU techniques, to better evaluate the content and the semantics of the given answers~\cite{Namazifar2021}.
This approach can also be used to identify weaknesses of the models and take appropriate countermeasures. This will be also beneficial to examine and control the relevance of spelling errors or meaningless sentences made of specific keywords to the computation of the answer score.

\section{Conclusions}
\label{sec:conclusions}
We reviewed the recent progress in the field of Automated Short Answer Grading (ASAG) and provided an overview of the advancements made and results achieved using deep learning techniques. We added to previous literature analyses by identifying the key features and architectural choices that impacted the performance of ASAG systems in the era of Deep Learning. We linked the methodological improvements to the results that recent methods achieved on benchmark data sets. 

This survey provides a taxonomy of methods, from classical Machine Learning to Deep Learning approaches, and research trends and outlook. Deep Learning architectures for natural language processing, adapted to ASAG tasks by means of transfer learning and domain adaptation techniques, are not sufficient to deal with the challenges and requirements that this field presents. Deep learning approaches alone show difficulty in effectively catch the semantics of short answers for consistent comparison with reference answers. To compensate for this, an ensemble of classifiers, stacked models, and  especially hybrid models that combine feature engineering with deep representation learning were developed. The embedding capabilities of Deep Learning models and the attention-based analysis of Transformers have been shown to be complementary to previously developed lexical, syntactic, and semantic features, to strengthen the performance of ASAG systems. However, to stimulate further developments, a uniform basis for benchmarking methods is necessary, e.g. designing a comprehensive benchmark data set.

\appendix
\bibliographystyle{ACM-Reference-Format}
\bibliography{sample-bibliography}


\begin{thebibliography}{84}


\ifx \showCODEN    \undefined \def \showCODEN     #1{\unskip}     \fi
\ifx \showDOI      \undefined \def \showDOI       #1{#1}\fi
\ifx \showISBNx    \undefined \def \showISBNx     #1{\unskip}     \fi
\ifx \showISBNxiii \undefined \def \showISBNxiii  #1{\unskip}     \fi
\ifx \showISSN     \undefined \def \showISSN      #1{\unskip}     \fi
\ifx \showLCCN     \undefined \def \showLCCN      #1{\unskip}     \fi
\ifx \shownote     \undefined \def \shownote      #1{#1}          \fi
\ifx \showarticletitle \undefined \def \showarticletitle #1{#1}   \fi
\ifx \showURL      \undefined \def \showURL       {\relax}        \fi
\providecommand\bibfield[2]{#2}
\providecommand\bibinfo[2]{#2}
\providecommand\natexlab[1]{#1}
\providecommand\showeprint[2][]{arXiv:#2}

\bibitem[\protect\citeauthoryear{Blessing, Azeta, Misra, Chigozie, and
  Ahuja}{Blessing et~al\mbox{.}}{2021}]%
        {Blessing2019AML}
\bibfield{author}{\bibinfo{person}{Guembe Blessing}, \bibinfo{person}{Ambrose
  Azeta}, \bibinfo{person}{Sanjay Misra}, \bibinfo{person}{Felix Chigozie},
  {and} \bibinfo{person}{Ravin Ahuja}.} \bibinfo{year}{2021}\natexlab{}.
\newblock \showarticletitle{A Machine Learning Prediction of Automatic Text
  Based Assessment for Open and Distance Learning: A Review}. In
  \bibinfo{booktitle}{\emph{Innovations in Bio-Inspired Computing and
  Applications}}. \bibinfo{publisher}{Springer International Publishing},
  \bibinfo{address}{Cham}, \bibinfo{pages}{369--380}.
\newblock
\showISBNx{978-3-030-49339-4}


\bibitem[\protect\citeauthoryear{Burrows, Gurevych, and Stein}{Burrows
  et~al\mbox{.}}{2015}]%
        {Burrows2014}
\bibfield{author}{\bibinfo{person}{Steven Burrows}, \bibinfo{person}{Iryna
  Gurevych}, {and} \bibinfo{person}{Benno Stein}.}
  \bibinfo{year}{2015}\natexlab{}.
\newblock \showarticletitle{The Eras and Trends of Automatic Short Answer
  Grading}.
\newblock \bibinfo{journal}{\emph{International Journal of Artificial
  Intelligence in Education}} \bibinfo{volume}{25}, \bibinfo{number}{1}
  (\bibinfo{date}{01 Mar} \bibinfo{year}{2015}), \bibinfo{pages}{60--117}.
\newblock
\showISSN{1560-4306}
\urldef\tempurl%
\url{https://doi.org/10.1007/s40593-014-0026-8}
\showDOI{\tempurl}


\bibitem[\protect\citeauthoryear{Cai}{Cai}{2019}]%
        {Cai2019}
\bibfield{author}{\bibinfo{person}{Changzhi Cai}.}
  \bibinfo{year}{2019}\natexlab{}.
\newblock \showarticletitle{Automatic Essay Scoring with Recurrent Neural
  Network}. In \bibinfo{booktitle}{\emph{Proceedings of the 3rd International
  Conference on High Performance Compilation, Computing and Communications}}.
  \bibinfo{publisher}{Association for Computing Machinery},
  \bibinfo{address}{New York, NY, USA}, \bibinfo{pages}{1–7}.
\newblock
\showISBNx{9781450366380}
\urldef\tempurl%
\url{https://doi.org/10.1145/3318265.3318296}
\showDOI{\tempurl}


\bibitem[\protect\citeauthoryear{Camus and Filighera}{Camus and
  Filighera}{2020}]%
        {Camus2020InvestigatingTF}
\bibfield{author}{\bibinfo{person}{Leon Camus} {and} \bibinfo{person}{Anna
  Filighera}.} \bibinfo{year}{2020}\natexlab{}.
\newblock \showarticletitle{Investigating Transformers for Automatic Short
  Answer Grading}. In \bibinfo{booktitle}{\emph{Artificial Intelligence in
  Education}}, \bibfield{editor}{\bibinfo{person}{Ig~Ibert Bittencourt},
  \bibinfo{person}{Mutlu Cukurova}, \bibinfo{person}{Kasia Muldner},
  \bibinfo{person}{Rose Luckin}, {and} \bibinfo{person}{Eva Mill{\'a}n}}
  (Eds.). \bibinfo{publisher}{Springer International Publishing},
  \bibinfo{address}{Cham}, \bibinfo{pages}{43--48}.
\newblock
\showISBNx{978-3-030-52240-7}


\bibitem[\protect\citeauthoryear{Cheng, Dong, and Lapata}{Cheng
  et~al\mbox{.}}{2016}]%
        {Cheng2016}
\bibfield{author}{\bibinfo{person}{Jianpeng Cheng}, \bibinfo{person}{Li Dong},
  {and} \bibinfo{person}{Mirella Lapata}.} \bibinfo{year}{2016}\natexlab{}.
\newblock \showarticletitle{Long Short-Term Memory-Networks for Machine
  Reading}. In \bibinfo{booktitle}{\emph{Proceedings of the 2016 Conference on
  Empirical Methods in Natural Language Processing}}.
  \bibinfo{publisher}{Association for Computational Linguistics},
  \bibinfo{address}{Austin, Texas}, \bibinfo{pages}{551--561}.
\newblock
\urldef\tempurl%
\url{https://doi.org/10.18653/v1/D16-1053}
\showDOI{\tempurl}


\bibitem[\protect\citeauthoryear{Condor}{Condor}{2020}]%
        {Condor2020ExploringAS}
\bibfield{author}{\bibinfo{person}{Aubrey Condor}.}
  \bibinfo{year}{2020}\natexlab{}.
\newblock \showarticletitle{Exploring Automatic Short Answer Grading as a Tool
  to Assist in Human Rating}.
\newblock \bibinfo{journal}{\emph{Artificial Intelligence in Education}}
  \bibinfo{volume}{12164} (\bibinfo{year}{2020}), \bibinfo{pages}{74 -- 79}.
\newblock


\bibitem[\protect\citeauthoryear{Conneau, Kiela, Schwenk, Barrault, and
  Bordes}{Conneau et~al\mbox{.}}{2017}]%
        {conneau-etal-2017-supervised}
\bibfield{author}{\bibinfo{person}{Alexis Conneau}, \bibinfo{person}{Douwe
  Kiela}, \bibinfo{person}{Holger Schwenk}, \bibinfo{person}{Lo{\"\i}c
  Barrault}, {and} \bibinfo{person}{Antoine Bordes}.}
  \bibinfo{year}{2017}\natexlab{}.
\newblock \showarticletitle{Supervised Learning of Universal Sentence
  Representations from Natural Language Inference Data}. In
  \bibinfo{booktitle}{\emph{Proceedings of the 2017 Conference on Empirical
  Methods in Natural Language Processing}}. \bibinfo{publisher}{Association for
  Computational Linguistics}, \bibinfo{address}{Copenhagen, Denmark},
  \bibinfo{pages}{670--680}.
\newblock
\urldef\tempurl%
\url{https://doi.org/10.18653/v1/D17-1070}
\showDOI{\tempurl}


\bibitem[\protect\citeauthoryear{{De Mulder}, Bethard, and Moens}{{De Mulder}
  et~al\mbox{.}}{2015}]%
        {DEMULDER201561}
\bibfield{author}{\bibinfo{person}{Wim {De Mulder}}, \bibinfo{person}{Steven
  Bethard}, {and} \bibinfo{person}{Marie-Francine Moens}.}
  \bibinfo{year}{2015}\natexlab{}.
\newblock \showarticletitle{A survey on the application of recurrent neural
  networks to statistical language modeling}.
\newblock \bibinfo{journal}{\emph{Computer Speech \& Language}}
  \bibinfo{volume}{30}, \bibinfo{number}{1} (\bibinfo{year}{2015}),
  \bibinfo{pages}{61--98}.
\newblock
\showISSN{0885-2308}
\urldef\tempurl%
\url{https://doi.org/10.1016/j.csl.2014.09.005}
\showDOI{\tempurl}


\bibitem[\protect\citeauthoryear{Devlin, Chang, Lee, and Toutanova}{Devlin
  et~al\mbox{.}}{2019}]%
        {Devlin2019BERTPO}
\bibfield{author}{\bibinfo{person}{Jacob Devlin}, \bibinfo{person}{Ming{-}Wei
  Chang}, \bibinfo{person}{Kenton Lee}, {and} \bibinfo{person}{Kristina
  Toutanova}.} \bibinfo{year}{2019}\natexlab{}.
\newblock \showarticletitle{{BERT:} Pre-training of Deep Bidirectional
  Transformers for Language Understanding}. In
  \bibinfo{booktitle}{\emph{Proceedings of the 2019 Conference of the North
  American Chapter of the Association for Computational Linguistics: Human
  Language Technologies, {NAACL-HLT} 2019, Minneapolis, MN, USA, June 2-7,
  2019, Volume 1 (Long and Short Papers)}},
  \bibfield{editor}{\bibinfo{person}{Jill Burstein}, \bibinfo{person}{Christy
  Doran}, {and} \bibinfo{person}{Thamar Solorio}} (Eds.).
  \bibinfo{publisher}{Association for Computational Linguistics},
  \bibinfo{pages}{4171--4186}.
\newblock
\urldef\tempurl%
\url{https://doi.org/10.18653/v1/n19-1423}
\showDOI{\tempurl}


\bibitem[\protect\citeauthoryear{Dhamecha, Marvaniya, Saha, Sindhgatta, and
  Sengupta}{Dhamecha et~al\mbox{.}}{2018}]%
        {LSdataset2018}
\bibfield{author}{\bibinfo{person}{Tejas~I. Dhamecha}, \bibinfo{person}{Smit
  Marvaniya}, \bibinfo{person}{Swarnadeep Saha}, \bibinfo{person}{Renuka
  Sindhgatta}, {and} \bibinfo{person}{Bikram Sengupta}.}
  \bibinfo{year}{2018}\natexlab{}.
\newblock \showarticletitle{Balancing Human Efforts and Performance of Student
  Response Analyzer in Dialog-Based Tutors}. In
  \bibinfo{booktitle}{\emph{Artificial Intelligence in Education}},
  \bibfield{editor}{\bibinfo{person}{Carolyn Penstein~Ros{\'e}},
  \bibinfo{person}{Roberto Mart{\'i}nez-Maldonado}, \bibinfo{person}{H.~Ulrich
  Hoppe}, \bibinfo{person}{Rose Luckin}, \bibinfo{person}{Manolis Mavrikis},
  \bibinfo{person}{Kaska Porayska-Pomsta}, \bibinfo{person}{Bruce McLaren},
  {and} \bibinfo{person}{Benedict du~Boulay}} (Eds.).
  \bibinfo{publisher}{Springer International Publishing},
  \bibinfo{address}{Cham}, \bibinfo{pages}{70--85}.
\newblock
\showISBNx{978-3-319-93843-1}


\bibitem[\protect\citeauthoryear{Dyer, Kuncoro, Ballesteros, and Smith}{Dyer
  et~al\mbox{.}}{2016}]%
        {dyer-etal-2016-recurrent}
\bibfield{author}{\bibinfo{person}{Chris Dyer}, \bibinfo{person}{Adhiguna
  Kuncoro}, \bibinfo{person}{Miguel Ballesteros}, {and}
  \bibinfo{person}{Noah~A. Smith}.} \bibinfo{year}{2016}\natexlab{}.
\newblock \showarticletitle{Recurrent Neural Network Grammars}. In
  \bibinfo{booktitle}{\emph{Proceedings of the 2016 Conference of the North
  {A}merican Chapter of the Association for Computational Linguistics: Human
  Language Technologies}}. \bibinfo{publisher}{Association for Computational
  Linguistics}, \bibinfo{address}{San Diego, California},
  \bibinfo{pages}{199--209}.
\newblock
\urldef\tempurl%
\url{https://doi.org/10.18653/v1/N16-1024}
\showDOI{\tempurl}


\bibitem[\protect\citeauthoryear{Dzikovska, Nielsen, Brew, Leacock,
  Giampiccolo, Bentivogli, Clark, Dagan, and Dang}{Dzikovska
  et~al\mbox{.}}{2013}]%
        {Dzikovska2013SemEval2013T7}
\bibfield{author}{\bibinfo{person}{Myroslava Dzikovska},
  \bibinfo{person}{Rodney Nielsen}, \bibinfo{person}{Chris Brew},
  \bibinfo{person}{Claudia Leacock}, \bibinfo{person}{Danilo Giampiccolo},
  \bibinfo{person}{Luisa Bentivogli}, \bibinfo{person}{Peter Clark},
  \bibinfo{person}{Ido Dagan}, {and} \bibinfo{person}{Hoa~Trang Dang}.}
  \bibinfo{year}{2013}\natexlab{}.
\newblock \showarticletitle{{S}em{E}val-2013 Task 7: The Joint Student Response
  Analysis and 8th Recognizing Textual Entailment Challenge}. In
  \bibinfo{booktitle}{\emph{Second Joint Conference on Lexical and
  Computational Semantics (*{SEM}), Volume 2: Proceedings of the Seventh
  International Workshop on Semantic Evaluation ({S}em{E}val 2013)}}.
  \bibinfo{publisher}{Association for Computational Linguistics},
  \bibinfo{address}{Atlanta, Georgia, USA}, \bibinfo{pages}{263--274}.
\newblock
\urldef\tempurl%
\url{https://aclanthology.org/S13-2045}
\showURL{%
\tempurl}


\bibitem[\protect\citeauthoryear{Dzikovska, Nielsen, and Brew}{Dzikovska
  et~al\mbox{.}}{2012}]%
        {Dzikovska2012TowardsET}
\bibfield{author}{\bibinfo{person}{Myroslava~O. Dzikovska},
  \bibinfo{person}{Rodney~D. Nielsen}, {and} \bibinfo{person}{Chris Brew}.}
  \bibinfo{year}{2012}\natexlab{}.
\newblock \showarticletitle{Towards Effective Tutorial Feedback for Explanation
  Questions: A Dataset and Baselines}. In \bibinfo{booktitle}{\emph{Proceedings
  of the 2012 Conference of the North {A}merican Chapter of the Association for
  Computational Linguistics: Human Language Technologies}}.
  \bibinfo{publisher}{Association for Computational Linguistics},
  \bibinfo{address}{Montr{\'e}al, Canada}, \bibinfo{pages}{200--210}.
\newblock
\urldef\tempurl%
\url{https://aclanthology.org/N12-1021}
\showURL{%
\tempurl}


\bibitem[\protect\citeauthoryear{Fellbaum}{Fellbaum}{2000}]%
        {Fellbaum2000WordNetA}
\bibfield{author}{\bibinfo{person}{C. Fellbaum}.}
  \bibinfo{year}{2000}\natexlab{}.
\newblock \showarticletitle{WordNet : an electronic lexical database}.
\newblock \bibinfo{journal}{\emph{Language}}  \bibinfo{volume}{76}
  (\bibinfo{year}{2000}), \bibinfo{pages}{706}.
\newblock


\bibitem[\protect\citeauthoryear{Gabrilovich and Markovitch}{Gabrilovich and
  Markovitch}{2007a}]%
        {Gabrilovich2007ESA}
\bibfield{author}{\bibinfo{person}{Evgeniy Gabrilovich} {and}
  \bibinfo{person}{Shaul Markovitch}.} \bibinfo{year}{2007}\natexlab{a}.
\newblock \showarticletitle{Computing Semantic Relatedness Using
  Wikipedia-Based Explicit Semantic Analysis}. In
  \bibinfo{booktitle}{\emph{Proceedings of the 20th International Joint
  Conference on Artifical Intelligence}} \emph{(\bibinfo{series}{IJCAI'07})}.
  \bibinfo{publisher}{Morgan Kaufmann Publishers Inc.}, \bibinfo{address}{San
  Francisco, CA, USA}, \bibinfo{pages}{1606–1611}.
\newblock


\bibitem[\protect\citeauthoryear{Gabrilovich and Markovitch}{Gabrilovich and
  Markovitch}{2007b}]%
        {Gabrilovich2007ComputingSR}
\bibfield{author}{\bibinfo{person}{Evgeniy Gabrilovich} {and}
  \bibinfo{person}{Shaul Markovitch}.} \bibinfo{year}{2007}\natexlab{b}.
\newblock \showarticletitle{Computing Semantic Relatedness Using
  Wikipedia-Based Explicit Semantic Analysis}. In
  \bibinfo{booktitle}{\emph{Proceedings of the 20th International Joint
  Conference on Artifical Intelligence}} \emph{(\bibinfo{series}{IJCAI'07})}.
  \bibinfo{publisher}{Morgan Kaufmann Publishers Inc.}, \bibinfo{address}{San
  Francisco, CA, USA}, \bibinfo{pages}{1606–1611}.
\newblock


\bibitem[\protect\citeauthoryear{Gaddipati, Nair, and Pl{\"o}ger}{Gaddipati
  et~al\mbox{.}}{2020}]%
        {Gaddipati2020ComparativeEO}
\bibfield{author}{\bibinfo{person}{Sasi~Kiran Gaddipati},
  \bibinfo{person}{Deebul Nair}, {and} \bibinfo{person}{P. Pl{\"o}ger}.}
  \bibinfo{year}{2020}\natexlab{}.
\newblock \showarticletitle{Comparative Evaluation of Pretrained Transfer
  Learning Models on Automatic Short Answer Grading}.
\newblock \bibinfo{journal}{\emph{ArXiv}}  \bibinfo{volume}{abs/2009.01303}
  (\bibinfo{year}{2020}).
\newblock


\bibitem[\protect\citeauthoryear{Galassi, Lippi, and Torroni}{Galassi
  et~al\mbox{.}}{2020}]%
        {Galassi2020}
\bibfield{author}{\bibinfo{person}{Andrea Galassi}, \bibinfo{person}{Marco
  Lippi}, {and} \bibinfo{person}{Paolo Torroni}.}
  \bibinfo{year}{2020}\natexlab{}.
\newblock \showarticletitle{Attention in Natural Language Processing}.
\newblock \bibinfo{journal}{\emph{IEEE Transactions on Neural Networks and
  Learning Systems}} (\bibinfo{year}{2020}), \bibinfo{pages}{1--18}.
\newblock
\urldef\tempurl%
\url{https://doi.org/10.1109/TNNLS.2020.3019893}
\showDOI{\tempurl}


\bibitem[\protect\citeauthoryear{Galhardi and Brancher}{Galhardi and
  Brancher}{2018}]%
        {Galhardi2018MachineLA}
\bibfield{author}{\bibinfo{person}{Lucas~Busatta Galhardi} {and}
  \bibinfo{person}{Jacques~Du{\'i}lio Brancher}.}
  \bibinfo{year}{2018}\natexlab{}.
\newblock \showarticletitle{Machine Learning Approach for Automatic Short
  Answer Grading: A Systematic Review}. In \bibinfo{booktitle}{\emph{Advances
  in Artificial Intelligence - IBERAMIA 2018}},
  \bibfield{editor}{\bibinfo{person}{Guillermo~R. Simari},
  \bibinfo{person}{Eduardo Ferm{\'e}}, \bibinfo{person}{Flabio
  Guti{\'e}rrez~Segura}, {and} \bibinfo{person}{Jos{\'e}~Antonio
  Rodr{\'i}guez~Melquiades}} (Eds.). \bibinfo{publisher}{Springer International
  Publishing}, \bibinfo{address}{Cham}, \bibinfo{pages}{380--391}.
\newblock
\showISBNx{978-3-030-03928-8}


\bibitem[\protect\citeauthoryear{Galhardi, Senefonte, de~Souza, and
  Brancher}{Galhardi et~al\mbox{.}}{2018}]%
        {Galhardi2018ExploringDF}
\bibfield{author}{\bibinfo{person}{Lucas~B. Galhardi}, \bibinfo{person}{Helen
  Senefonte}, \bibinfo{person}{Rodrigo de Souza}, {and}
  \bibinfo{person}{Jacques Brancher}.} \bibinfo{year}{2018}\natexlab{}.
\newblock \showarticletitle{Exploring Distinct Features for Automatic Short
  Answer Grading}. In \bibinfo{booktitle}{\emph{Anais do XV Encontro Nacional
  de Inteligência Artificial e Computacional}}. \bibinfo{publisher}{SBC},
  \bibinfo{address}{Porto Alegre, RS, Brasil}, \bibinfo{pages}{1--12}.
\newblock
\showISSN{0000-0000}
\urldef\tempurl%
\url{https://doi.org/10.5753/eniac.2018.4399}
\showDOI{\tempurl}


\bibitem[\protect\citeauthoryear{Gomaa and Fahmy}{Gomaa and Fahmy}{2020}]%
        {Gomaa2019Ans2vecAS}
\bibfield{author}{\bibinfo{person}{Wael~Hassan Gomaa} {and}
  \bibinfo{person}{Aly~Aly Fahmy}.} \bibinfo{year}{2020}\natexlab{}.
\newblock \showarticletitle{Ans2vec: A Scoring System for Short Answers}. In
  \bibinfo{booktitle}{\emph{The International Conference on Advanced Machine
  Learning Technologies and Applications (AMLTA2019)}},
  \bibfield{editor}{\bibinfo{person}{Aboul~Ella Hassanien},
  \bibinfo{person}{Ahmad~Taher Azar}, \bibinfo{person}{Tarek Gaber},
  \bibinfo{person}{Roheet Bhatnagar}, {and} \bibinfo{person}{Mohamed F.~Tolba}}
  (Eds.). \bibinfo{publisher}{Springer International Publishing},
  \bibinfo{address}{Cham}, \bibinfo{pages}{586--595}.
\newblock
\showISBNx{978-3-030-14118-9}


\bibitem[\protect\citeauthoryear{Guidotti, Monreale, Ruggieri, Turini,
  Giannotti, and Pedreschi}{Guidotti et~al\mbox{.}}{2018}]%
        {Guidotti19}
\bibfield{author}{\bibinfo{person}{Riccardo Guidotti}, \bibinfo{person}{Anna
  Monreale}, \bibinfo{person}{Salvatore Ruggieri}, \bibinfo{person}{Franco
  Turini}, \bibinfo{person}{Fosca Giannotti}, {and} \bibinfo{person}{Dino
  Pedreschi}.} \bibinfo{year}{2018}\natexlab{}.
\newblock \showarticletitle{A Survey of Methods for Explaining Black Box
  Models}.
\newblock \bibinfo{journal}{\emph{ACM Comput. Surv.}} \bibinfo{volume}{51},
  \bibinfo{number}{5}, Article \bibinfo{articleno}{93} (\bibinfo{year}{2018}),
  \bibinfo{numpages}{42}~pages.
\newblock
\showISSN{0360-0300}
\urldef\tempurl%
\url{https://doi.org/10.1145/3236009}
\showDOI{\tempurl}


\bibitem[\protect\citeauthoryear{Heilman and Madnani}{Heilman and
  Madnani}{2012}]%
        {Heilman2012ETSDE}
\bibfield{author}{\bibinfo{person}{Michael Heilman} {and}
  \bibinfo{person}{Nitin Madnani}.} \bibinfo{year}{2012}\natexlab{}.
\newblock \showarticletitle{{ETS}: Discriminative Edit Models for Paraphrase
  Scoring}. In \bibinfo{booktitle}{\emph{*{SEM} 2012: The First Joint
  Conference on Lexical and Computational Semantics {--} Volume 1: Proceedings
  of the main conference and the shared task, and Volume 2: Proceedings of the
  Sixth International Workshop on Semantic Evaluation ({S}em{E}val 2012)}}.
  \bibinfo{publisher}{Association for Computational Linguistics},
  \bibinfo{address}{Montr{\'e}al, Canada}, \bibinfo{pages}{529--535}.
\newblock
\urldef\tempurl%
\url{https://aclanthology.org/S12-1076}
\showURL{%
\tempurl}


\bibitem[\protect\citeauthoryear{Heilman and Madnani}{Heilman and
  Madnani}{2013}]%
        {Heilman2013ETSDA}
\bibfield{author}{\bibinfo{person}{Michael Heilman} {and}
  \bibinfo{person}{Nitin Madnani}.} \bibinfo{year}{2013}\natexlab{}.
\newblock \showarticletitle{{ETS}: Domain Adaptation and Stacking for Short
  Answer Scoring}. In \bibinfo{booktitle}{\emph{Second Joint Conference on
  Lexical and Computational Semantics (*{SEM}), Volume 2: Proceedings of the
  Seventh International Workshop on Semantic Evaluation ({S}em{E}val 2013)}}.
  \bibinfo{publisher}{Association for Computational Linguistics},
  \bibinfo{address}{Atlanta, Georgia, USA}, \bibinfo{pages}{275--279}.
\newblock
\urldef\tempurl%
\url{https://aclanthology.org/S13-2046}
\showURL{%
\tempurl}


\bibitem[\protect\citeauthoryear{Hightower}{Hightower}{2020}]%
        {Hightower}
\bibfield{author}{\bibinfo{person}{Mallory Hightower}.}
  \bibinfo{year}{2020}\natexlab{}.
\newblock \bibinfo{title}{High-Level History of NLP Models}.
\newblock
\newblock
\urldef\tempurl%
\url{https://towardsdatascience.com/high-level-history-of-nlp-models-bc8c8b142ef7}
\showURL{%
\tempurl}


\bibitem[\protect\citeauthoryear{Howard and Ruder}{Howard and Ruder}{2018}]%
        {Howard2018UniversalLM}
\bibfield{author}{\bibinfo{person}{Jeremy Howard} {and}
  \bibinfo{person}{Sebastian Ruder}.} \bibinfo{year}{2018}\natexlab{}.
\newblock \showarticletitle{Universal Language Model Fine-tuning for Text
  Classification}. In \bibinfo{booktitle}{\emph{Proceedings of the 56th Annual
  Meeting of the Association for Computational Linguistics (Volume 1: Long
  Papers)}}. \bibinfo{publisher}{Association for Computational Linguistics},
  \bibinfo{address}{Melbourne, Australia}, \bibinfo{pages}{328--339}.
\newblock
\urldef\tempurl%
\url{https://doi.org/10.18653/v1/P18-1031}
\showDOI{\tempurl}


\bibitem[\protect\citeauthoryear{Jimenez, Becerra, and Gelbukh}{Jimenez
  et~al\mbox{.}}{2013}]%
        {Jimnez2013SOFTCARDINALITYHT}
\bibfield{author}{\bibinfo{person}{Sergio Jimenez}, \bibinfo{person}{Claudia
  Becerra}, {and} \bibinfo{person}{Alexander Gelbukh}.}
  \bibinfo{year}{2013}\natexlab{}.
\newblock \showarticletitle{{SOFTCARDINALITY}: Hierarchical Text Overlap for
  Student Response Analysis}. In \bibinfo{booktitle}{\emph{Second Joint
  Conference on Lexical and Computational Semantics (*{SEM}), Volume 2:
  Proceedings of the Seventh International Workshop on Semantic Evaluation
  ({S}em{E}val 2013)}}. \bibinfo{publisher}{Association for Computational
  Linguistics}, \bibinfo{address}{Atlanta, Georgia, USA},
  \bibinfo{pages}{280--284}.
\newblock
\urldef\tempurl%
\url{https://aclanthology.org/S13-2047}
\showURL{%
\tempurl}


\bibitem[\protect\citeauthoryear{Kaggle}{Kaggle}{[n. d.]}]%
        {asap-sas}
\bibfield{author}{\bibinfo{person}{Kaggle}.} \bibinfo{year}{[n.
  d.]}\natexlab{}.
\newblock \bibinfo{title}{The Hewlett Foundation: Short Answer Scoring}.
\newblock
\newblock
\urldef\tempurl%
\url{https://www.kaggle.com/c/asap-sas/}
\showURL{%
\tempurl}


\bibitem[\protect\citeauthoryear{Kaur and Hornof}{Kaur and Hornof}{2005}]%
        {Kaur2005ACO}
\bibfield{author}{\bibinfo{person}{Ishwinder Kaur} {and}
  \bibinfo{person}{Anthony~J. Hornof}.} \bibinfo{year}{2005}\natexlab{}.
\newblock \bibinfo{booktitle}{\emph{A Comparison of LSA, WordNet and PMI-IR for
  Predicting User Click Behavior}}.
\newblock \bibinfo{publisher}{Association for Computing Machinery},
  \bibinfo{address}{New York, NY, USA}, \bibinfo{pages}{51–60}.
\newblock
\showISBNx{1581139985}
\urldef\tempurl%
\url{https://doi.org/10.1145/1054972.1054980}
\showURL{%
\tempurl}


\bibitem[\protect\citeauthoryear{Kouylekov, Dini, Bosca, and
  Trevisan}{Kouylekov et~al\mbox{.}}{2013}]%
        {Kouylekov2013CeliEA}
\bibfield{author}{\bibinfo{person}{Milen Kouylekov}, \bibinfo{person}{Luca
  Dini}, \bibinfo{person}{Alessio Bosca}, {and} \bibinfo{person}{Marco
  Trevisan}.} \bibinfo{year}{2013}\natexlab{}.
\newblock \showarticletitle{{C}eli: {EDITS} and Generic Text Pair
  Classification}. In \bibinfo{booktitle}{\emph{Second Joint Conference on
  Lexical and Computational Semantics (*{SEM}), Volume 2: Proceedings of the
  Seventh International Workshop on Semantic Evaluation ({S}em{E}val 2013)}}.
  \bibinfo{publisher}{Association for Computational Linguistics},
  \bibinfo{address}{Atlanta, Georgia, USA}, \bibinfo{pages}{592--597}.
\newblock
\urldef\tempurl%
\url{https://aclanthology.org/S13-2099}
\showURL{%
\tempurl}


\bibitem[\protect\citeauthoryear{Kumar, Chakrabarti, and Roy}{Kumar
  et~al\mbox{.}}{2017}]%
        {Kumar2017EarthMD}
\bibfield{author}{\bibinfo{person}{Sachin Kumar}, \bibinfo{person}{Soumen
  Chakrabarti}, {and} \bibinfo{person}{Shourya Roy}.}
  \bibinfo{year}{2017}\natexlab{}.
\newblock \showarticletitle{Earth Mover's Distance Pooling over Siamese LSTMs
  for Automatic Short Answer Grading}. In \bibinfo{booktitle}{\emph{Proceedings
  of the Twenty-Sixth International Joint Conference on Artificial
  Intelligence, {IJCAI-17}}}. \bibinfo{pages}{2046--2052}.
\newblock
\urldef\tempurl%
\url{https://doi.org/10.24963/ijcai.2017/284}
\showDOI{\tempurl}


\bibitem[\protect\citeauthoryear{Kumar, Aggarwal, Mahata, Shah, Kumaraguru, and
  Zimmermann}{Kumar et~al\mbox{.}}{2019}]%
        {Kumar2019GetIS}
\bibfield{author}{\bibinfo{person}{Yaman Kumar}, \bibinfo{person}{Swati
  Aggarwal}, \bibinfo{person}{Debanjan Mahata}, \bibinfo{person}{Rajiv~Ratn
  Shah}, \bibinfo{person}{Ponnurangam Kumaraguru}, {and} \bibinfo{person}{Roger
  Zimmermann}.} \bibinfo{year}{2019}\natexlab{}.
\newblock \showarticletitle{Get IT Scored Using AutoSAS - An Automated System
  for Scoring Short Answers}. In \bibinfo{booktitle}{\emph{AAAI}}.
\newblock


\bibitem[\protect\citeauthoryear{Lan, Chen, Goodman, Gimpel, Sharma, and
  Soricut}{Lan et~al\mbox{.}}{2020}]%
        {Lan2020ALBERTAL}
\bibfield{author}{\bibinfo{person}{Zhenzhong Lan}, \bibinfo{person}{Mingda
  Chen}, \bibinfo{person}{Sebastian Goodman}, \bibinfo{person}{Kevin Gimpel},
  \bibinfo{person}{Piyush Sharma}, {and} \bibinfo{person}{Radu Soricut}.}
  \bibinfo{year}{2020}\natexlab{}.
\newblock \showarticletitle{ALBERT: A Lite BERT for Self-supervised Learning of
  Language Representations}.
\newblock  (\bibinfo{year}{2020}).
\newblock
\urldef\tempurl%
\url{https://openreview.net/forum?id=H1eA7AEtvS}
\showURL{%
\tempurl}


\bibitem[\protect\citeauthoryear{Landauer, Foltz, and Laham}{Landauer
  et~al\mbox{.}}{1998}]%
        {Landauer1998LSA}
\bibfield{author}{\bibinfo{person}{Thomas~K Landauer},
  \bibinfo{person}{Peter~W. Foltz}, {and} \bibinfo{person}{Darrell Laham}.}
  \bibinfo{year}{1998}\natexlab{}.
\newblock \showarticletitle{An introduction to latent semantic analysis}.
\newblock \bibinfo{journal}{\emph{Discourse Processes}} \bibinfo{volume}{25},
  \bibinfo{number}{2-3} (\bibinfo{year}{1998}), \bibinfo{pages}{259--284}.
\newblock
\urldef\tempurl%
\url{https://doi.org/10.1080/01638539809545028}
\showDOI{\tempurl}


\bibitem[\protect\citeauthoryear{Levy, Zesch, Dagan, and Gurevych}{Levy
  et~al\mbox{.}}{2013}]%
        {Levy2013UKPBIUSA}
\bibfield{author}{\bibinfo{person}{Omer Levy}, \bibinfo{person}{Torsten Zesch},
  \bibinfo{person}{Ido Dagan}, {and} \bibinfo{person}{Iryna Gurevych}.}
  \bibinfo{year}{2013}\natexlab{}.
\newblock In \bibinfo{booktitle}{\emph{Second Joint Conference on Lexical and
  Computational Semantics (*{SEM}), Volume 2: Proceedings of the Seventh
  International Workshop on Semantic Evaluation ({S}em{E}val 2013)}}.
  \bibinfo{publisher}{Association for Computational Linguistics},
  \bibinfo{address}{Atlanta, Georgia, USA}, \bibinfo{pages}{285--289}.
\newblock


\bibitem[\protect\citeauthoryear{Liu, Ding, Wang, Tang, Huang, and Liu}{Liu
  et~al\mbox{.}}{2019a}]%
        {Liu2019AutomaticSA}
\bibfield{author}{\bibinfo{person}{Tianqiao Liu}, \bibinfo{person}{Wenbiao
  Ding}, \bibinfo{person}{Zhiwei Wang}, \bibinfo{person}{Jiliang Tang},
  \bibinfo{person}{Gale~Yan Huang}, {and} \bibinfo{person}{Zitao Liu}.}
  \bibinfo{year}{2019}\natexlab{a}.
\newblock \showarticletitle{Automatic Short Answer Grading via Multiway
  Attention Networks}.
\newblock \bibinfo{journal}{\emph{ArXiv}}  \bibinfo{volume}{abs/1909.10166}
  (\bibinfo{year}{2019}).
\newblock


\bibitem[\protect\citeauthoryear{Liu, He, Chen, and Gao}{Liu
  et~al\mbox{.}}{2019b}]%
        {Liu2019ImprovingMD}
\bibfield{author}{\bibinfo{person}{Xiaodong Liu}, \bibinfo{person}{Pengcheng
  He}, \bibinfo{person}{Weizhu Chen}, {and} \bibinfo{person}{Jianfeng Gao}.}
  \bibinfo{year}{2019}\natexlab{b}.
\newblock \showarticletitle{Improving Multi-Task Deep Neural Networks via
  Knowledge Distillation for Natural Language Understanding}.
\newblock \bibinfo{journal}{\emph{ArXiv}}  \bibinfo{volume}{abs/1904.09482}
  (\bibinfo{year}{2019}).
\newblock


\bibitem[\protect\citeauthoryear{Liu, Ott, Goyal, Du, Joshi, Chen, Levy, Lewis,
  Zettlemoyer, and Stoyanov}{Liu et~al\mbox{.}}{2019c}]%
        {Liu2019RoBERTaAR}
\bibfield{author}{\bibinfo{person}{Yinhan Liu}, \bibinfo{person}{Myle Ott},
  \bibinfo{person}{Naman Goyal}, \bibinfo{person}{Jingfei Du},
  \bibinfo{person}{Mandar Joshi}, \bibinfo{person}{Danqi Chen},
  \bibinfo{person}{Omer Levy}, \bibinfo{person}{Mike Lewis},
  \bibinfo{person}{Luke Zettlemoyer}, {and} \bibinfo{person}{Veselin
  Stoyanov}.} \bibinfo{year}{2019}\natexlab{c}.
\newblock \showarticletitle{RoBERTa: A Robustly Optimized BERT Pretraining
  Approach}.
\newblock  (\bibinfo{year}{2019}).
\newblock
\urldef\tempurl%
\url{http://arxiv.org/abs/1907.11692}
\showURL{%
\tempurl}
\newblock
\shownote{cite arxiv:1907.11692.}


\bibitem[\protect\citeauthoryear{Magooda, Zahran, Rashwan, Raafat, and
  Fayek}{Magooda et~al\mbox{.}}{2016}]%
        {Magooda2016}
\bibfield{author}{\bibinfo{person}{Ahmed~Ezzat Magooda},
  \bibinfo{person}{Mohamed~A. Zahran}, \bibinfo{person}{Mohsen~A. Rashwan},
  \bibinfo{person}{Hazem~M. Raafat}, {and} \bibinfo{person}{Magda~B. Fayek}.}
  \bibinfo{year}{2016}\natexlab{}.
\newblock \showarticletitle{Vector Based Techniques for Short Answer Grading}.
  In \bibinfo{booktitle}{\emph{Proceedings of the Twenty-Ninth International
  Florida Artificial Intelligence Research Society Conference, {FLAIRS} 2016,
  Key Largo, Florida, USA, May 16-18, 2016}},
  \bibfield{editor}{\bibinfo{person}{Zdravko Markov} {and}
  \bibinfo{person}{Ingrid Russell}} (Eds.). \bibinfo{publisher}{{AAAI} Press},
  \bibinfo{pages}{238--243}.
\newblock


\bibitem[\protect\citeauthoryear{Meurers, Ziai, Ott, and Bailey}{Meurers
  et~al\mbox{.}}{2011}]%
        {Meurers2011IntegratingPA}
\bibfield{author}{\bibinfo{person}{Detmar Meurers}, \bibinfo{person}{Ramon
  Ziai}, \bibinfo{person}{Niels Ott}, {and} \bibinfo{person}{Stacey~M Bailey}.}
  \bibinfo{year}{2011}\natexlab{}.
\newblock \showarticletitle{Integrating parallel analysis modules to evaluate
  the meaning of answers to reading comprehension questions}.
\newblock \bibinfo{journal}{\emph{International journal of continuing
  engineering education and life-long learning}}  \bibinfo{volume}{21}
  (\bibinfo{year}{2011}), \bibinfo{pages}{355--369}.
\newblock


\bibitem[\protect\citeauthoryear{Mikolov, Chen, Corrado, and Dean}{Mikolov
  et~al\mbox{.}}{2013}]%
        {Mikolov2013EfficientEO}
\bibfield{author}{\bibinfo{person}{Tomas Mikolov}, \bibinfo{person}{Kai Chen},
  \bibinfo{person}{G. Corrado}, {and} \bibinfo{person}{J. Dean}.}
  \bibinfo{year}{2013}\natexlab{}.
\newblock \showarticletitle{Efficient Estimation of Word Representations in
  Vector Space}. In \bibinfo{booktitle}{\emph{ICLR}}.
\newblock


\bibitem[\protect\citeauthoryear{Miller}{Miller}{1995}]%
        {Miller1995WordNet}
\bibfield{author}{\bibinfo{person}{George~A. Miller}.}
  \bibinfo{year}{1995}\natexlab{}.
\newblock \showarticletitle{WordNet: A Lexical Database for English}.
\newblock \bibinfo{journal}{\emph{Commun. ACM}} \bibinfo{volume}{38},
  \bibinfo{number}{11} (\bibinfo{date}{Nov.} \bibinfo{year}{1995}),
  \bibinfo{pages}{39–41}.
\newblock
\showISSN{0001-0782}
\urldef\tempurl%
\url{https://doi.org/10.1145/219717.219748}
\showDOI{\tempurl}


\bibitem[\protect\citeauthoryear{Mohler, Bunescu, and Mihalcea}{Mohler
  et~al\mbox{.}}{2011}]%
        {Mohler2011LearningTG}
\bibfield{author}{\bibinfo{person}{Michael Mohler}, \bibinfo{person}{Razvan
  Bunescu}, {and} \bibinfo{person}{Rada Mihalcea}.}
  \bibinfo{year}{2011}\natexlab{}.
\newblock \showarticletitle{Learning to Grade Short Answer Questions using
  Semantic Similarity Measures and Dependency Graph Alignments}. In
  \bibinfo{booktitle}{\emph{Proceedings of the 49th Annual Meeting of the
  Association for Computational Linguistics: Human Language Technologies}}.
  \bibinfo{publisher}{Association for Computational Linguistics},
  \bibinfo{address}{Portland, Oregon, USA}, \bibinfo{pages}{752--762}.
\newblock


\bibitem[\protect\citeauthoryear{Mohler and Mihalcea}{Mohler and
  Mihalcea}{2009}]%
        {Mohler2009}
\bibfield{author}{\bibinfo{person}{Michael Mohler} {and} \bibinfo{person}{Rada
  Mihalcea}.} \bibinfo{year}{2009}\natexlab{}.
\newblock \showarticletitle{Text-to-Text Semantic Similarity for Automatic
  Short Answer Grading}. In \bibinfo{booktitle}{\emph{Proceedings of the 12th
  Conference of the {E}uropean Chapter of the {ACL} ({EACL} 2009)}}.
  \bibinfo{publisher}{Association for Computational Linguistics},
  \bibinfo{address}{Athens, Greece}, \bibinfo{pages}{567--575}.
\newblock


\bibitem[\protect\citeauthoryear{Namazifar, Papangelis, Tur, and
  Hakkani-Tür}{Namazifar et~al\mbox{.}}{2021}]%
        {Namazifar2021}
\bibfield{author}{\bibinfo{person}{Mahdi Namazifar},
  \bibinfo{person}{Alexandros Papangelis}, \bibinfo{person}{Gokhan Tur}, {and}
  \bibinfo{person}{Dilek Hakkani-Tür}.} \bibinfo{year}{2021}\natexlab{}.
\newblock \showarticletitle{Language Model is all You Need: Natural Language
  Understanding as Question Answering}. In \bibinfo{booktitle}{\emph{ICASSP
  2021 - 2021 IEEE International Conference on Acoustics, Speech and Signal
  Processing (ICASSP)}}. \bibinfo{pages}{7803--7807}.
\newblock
\urldef\tempurl%
\url{https://doi.org/10.1109/ICASSP39728.2021.9413810}
\showDOI{\tempurl}


\bibitem[\protect\citeauthoryear{Navigli}{Navigli}{2018}]%
        {Navigli2018NLU}
\bibfield{author}{\bibinfo{person}{Roberto Navigli}.}
  \bibinfo{year}{2018}\natexlab{}.
\newblock \showarticletitle{Natural Language Understanding: Instructions for
  (Present and Future) Use}. In \bibinfo{booktitle}{\emph{Proceedings of the
  27th International Joint Conference on Artificial Intelligence}}
  \emph{(\bibinfo{series}{IJCAI'18})}. \bibinfo{publisher}{AAAI Press},
  \bibinfo{pages}{5697–5702}.
\newblock
\showISBNx{9780999241127}


\bibitem[\protect\citeauthoryear{Neterer and Guzide}{Neterer and
  Guzide}{2018}]%
        {Neterer2018DeepLI}
\bibfield{author}{\bibinfo{person}{Jacob~Russell Neterer} {and}
  \bibinfo{person}{Osman Guzide}.} \bibinfo{year}{2018}\natexlab{}.
\newblock \showarticletitle{Deep Learning in Natural Language Processing}.
\newblock \bibinfo{journal}{\emph{Proceedings of the West Virginia Academy of
  Science}} \bibinfo{volume}{90}, \bibinfo{number}{1}.
\newblock
\urldef\tempurl%
\url{https://pwvas.org/index.php/pwvas/article/view/339}
\showURL{%
\tempurl}


\bibitem[\protect\citeauthoryear{Ott, Ziai, Hahn, and Meurers}{Ott
  et~al\mbox{.}}{2013}]%
        {Ott2013CoMeTID}
\bibfield{author}{\bibinfo{person}{Niels Ott}, \bibinfo{person}{Ramon Ziai},
  \bibinfo{person}{Michael Hahn}, {and} \bibinfo{person}{Detmar Meurers}.}
  \bibinfo{year}{2013}\natexlab{}.
\newblock \showarticletitle{{C}o{M}e{T}: Integrating different levels of
  linguistic modeling for meaning assessment}. In
  \bibinfo{booktitle}{\emph{Second Joint Conference on Lexical and
  Computational Semantics (*{SEM}), Volume 2: Proceedings of the Seventh
  International Workshop on Semantic Evaluation ({S}em{E}val 2013)}}.
  \bibinfo{publisher}{Association for Computational Linguistics},
  \bibinfo{address}{Atlanta, Georgia, USA}, \bibinfo{pages}{608--616}.
\newblock
\urldef\tempurl%
\url{https://aclanthology.org/S13-2102}
\showURL{%
\tempurl}


\bibitem[\protect\citeauthoryear{Otter, Medina, and Kalita}{Otter
  et~al\mbox{.}}{2021}]%
        {Otter2020}
\bibfield{author}{\bibinfo{person}{Daniel~W. Otter}, \bibinfo{person}{Julian~R.
  Medina}, {and} \bibinfo{person}{Jugal~K. Kalita}.}
  \bibinfo{year}{2021}\natexlab{}.
\newblock \showarticletitle{A Survey of the Usages of Deep Learning for Natural
  Language Processing}.
\newblock \bibinfo{journal}{\emph{IEEE Transactions on Neural Networks and
  Learning Systems}} \bibinfo{volume}{32}, \bibinfo{number}{2}
  (\bibinfo{year}{2021}), \bibinfo{pages}{604--624}.
\newblock
\urldef\tempurl%
\url{https://doi.org/10.1109/TNNLS.2020.2979670}
\showDOI{\tempurl}


\bibitem[\protect\citeauthoryear{Pennington, Socher, and Manning}{Pennington
  et~al\mbox{.}}{2014}]%
        {Pennington2014GloveGV}
\bibfield{author}{\bibinfo{person}{Jeffrey Pennington}, \bibinfo{person}{R.
  Socher}, {and} \bibinfo{person}{Christopher~D. Manning}.}
  \bibinfo{year}{2014}\natexlab{}.
\newblock \showarticletitle{Glove: Global Vectors for Word Representation}. In
  \bibinfo{booktitle}{\emph{EMNLP}}.
\newblock


\bibitem[\protect\citeauthoryear{Peters, Neumann, Iyyer, Gardner, Clark, Lee,
  and Zettlemoyer}{Peters et~al\mbox{.}}{2018}]%
        {Peters2018DeepCW}
\bibfield{author}{\bibinfo{person}{Matthew~E. Peters}, \bibinfo{person}{Mark
  Neumann}, \bibinfo{person}{Mohit Iyyer}, \bibinfo{person}{Matt Gardner},
  \bibinfo{person}{Christopher Clark}, \bibinfo{person}{Kenton Lee}, {and}
  \bibinfo{person}{Luke Zettlemoyer}.} \bibinfo{year}{2018}\natexlab{}.
\newblock \showarticletitle{Deep Contextualized Word Representations}.
\newblock  (\bibinfo{date}{June} \bibinfo{year}{2018}),
  \bibinfo{pages}{2227--2237}.
\newblock
\urldef\tempurl%
\url{https://doi.org/10.18653/v1/N18-1202}
\showDOI{\tempurl}


\bibitem[\protect\citeauthoryear{Popovi{\'c}}{Popovi{\'c}}{2011}]%
        {Popovic2011MorphemesAP}
\bibfield{author}{\bibinfo{person}{Maja Popovi{\'c}}.}
  \bibinfo{year}{2011}\natexlab{}.
\newblock \showarticletitle{Morphemes and {POS} tags for n-gram based
  evaluation metrics}. In \bibinfo{booktitle}{\emph{Proceedings of the Sixth
  Workshop on Statistical Machine Translation}}.
  \bibinfo{publisher}{Association for Computational Linguistics},
  \bibinfo{address}{Edinburgh, Scotland}, \bibinfo{pages}{104--107}.
\newblock
\urldef\tempurl%
\url{https://aclanthology.org/W11-2110}
\showURL{%
\tempurl}


\bibitem[\protect\citeauthoryear{Pribadi, Adji, and Permanasari}{Pribadi
  et~al\mbox{.}}{2016}]%
        {Pribadi2016AutomatedSA}
\bibfield{author}{\bibinfo{person}{Feddy~Setio Pribadi},
  \bibinfo{person}{Teguh~Bharata Adji}, {and} \bibinfo{person}{Adhistya~Erna
  Permanasari}.} \bibinfo{year}{2016}\natexlab{}.
\newblock \showarticletitle{Automated Short Answer Scoring using Weighted
  Cosine Coefficient}.
\newblock  (\bibinfo{year}{2016}), \bibinfo{pages}{70--74}.
\newblock
\urldef\tempurl%
\url{https://doi.org/10.1109/IC3e.2016.8009042}
\showDOI{\tempurl}


\bibitem[\protect\citeauthoryear{Pribadi, Adji, Permanasari, Mulwinda, and
  Utomo}{Pribadi et~al\mbox{.}}{2017}]%
        {Pribadi2017AutomaticSA}
\bibfield{author}{\bibinfo{person}{Feddy~Setio Pribadi},
  \bibinfo{person}{Teguh~Bharata Adji}, \bibinfo{person}{Adhistya~Erna
  Permanasari}, \bibinfo{person}{Anggraini Mulwinda}, {and}
  \bibinfo{person}{Aryo~Baskoro Utomo}.} \bibinfo{year}{2017}\natexlab{}.
\newblock \showarticletitle{Automatic short answer scoring using words
  overlapping methods}.
\newblock \bibinfo{journal}{\emph{AIP Conference Proceedings}}
  \bibinfo{volume}{1818}, \bibinfo{number}{1}, \bibinfo{pages}{020042}.
\newblock
\urldef\tempurl%
\url{https://doi.org/10.1063/1.4976906}
\showDOI{\tempurl}


\bibitem[\protect\citeauthoryear{Qi, Wang, Dai, Li, and Di}{Qi
  et~al\mbox{.}}{2019}]%
        {Qi2019AttentionBasedHM}
\bibfield{author}{\bibinfo{person}{Hui Qi}, \bibinfo{person}{Yue Wang},
  \bibinfo{person}{Jinyu Dai}, \bibinfo{person}{Jinqing Li}, {and}
  \bibinfo{person}{Xiaoqiang Di}.} \bibinfo{year}{2019}\natexlab{}.
\newblock \showarticletitle{Attention-Based Hybrid Model for Automatic Short
  Answer Scoring}. In \bibinfo{booktitle}{\emph{Simulation Tools and
  Techniques}}, \bibfield{editor}{\bibinfo{person}{Houbing Song} {and}
  \bibinfo{person}{Dingde Jiang}} (Eds.). \bibinfo{publisher}{Springer
  International Publishing}, \bibinfo{address}{Cham},
  \bibinfo{pages}{385--394}.
\newblock
\showISBNx{978-3-030-32216-8}


\bibitem[\protect\citeauthoryear{Radford}{Radford}{2018}]%
        {Radford2018ImprovingLU}
\bibfield{author}{\bibinfo{person}{Alec Radford}.}
  \bibinfo{year}{2018}\natexlab{}.
\newblock \showarticletitle{Improving Language Understanding by Generative
  Pre-Training}.
\newblock


\bibitem[\protect\citeauthoryear{Raffel, Shazeer, Roberts, Lee, Narang, Matena,
  Zhou, Li, and Liu}{Raffel et~al\mbox{.}}{2019}]%
        {Raffel2019ExploringTL}
\bibfield{author}{\bibinfo{person}{Colin Raffel}, \bibinfo{person}{Noam
  Shazeer}, \bibinfo{person}{Adam Roberts}, \bibinfo{person}{Katherine Lee},
  \bibinfo{person}{Sharan Narang}, \bibinfo{person}{Michael Matena},
  \bibinfo{person}{Yanqi Zhou}, \bibinfo{person}{Wei Li}, {and}
  \bibinfo{person}{Peter~J. Liu}.} \bibinfo{year}{2019}\natexlab{}.
\newblock \showarticletitle{Exploring the Limits of Transfer Learning with a
  Unified Text-to-Text Transformer}.
\newblock \bibinfo{journal}{\emph{ArXiv}}  \bibinfo{volume}{abs/1910.10683}
  (\bibinfo{year}{2019}).
\newblock


\bibitem[\protect\citeauthoryear{Ramachandran, Cheng, and Foltz}{Ramachandran
  et~al\mbox{.}}{2015}]%
        {Ramachandran2015IdentifyingPF}
\bibfield{author}{\bibinfo{person}{Lakshmi Ramachandran}, \bibinfo{person}{Jian
  Cheng}, {and} \bibinfo{person}{Peter Foltz}.}
  \bibinfo{year}{2015}\natexlab{}.
\newblock \showarticletitle{Identifying Patterns For Short Answer Scoring Using
  Graph-based Lexico-Semantic Text Matching}. In
  \bibinfo{booktitle}{\emph{Proceedings of the Tenth Workshop on Innovative Use
  of {NLP} for Building Educational Applications}}.
  \bibinfo{publisher}{Association for Computational Linguistics},
  \bibinfo{address}{Denver, Colorado}, \bibinfo{pages}{97--106}.
\newblock
\urldef\tempurl%
\url{https://doi.org/10.3115/v1/W15-0612}
\showDOI{\tempurl}


\bibitem[\protect\citeauthoryear{Ratner, Hancock, Dunnmon, Sala, Pandey, and
  R{\'e}}{Ratner et~al\mbox{.}}{2019}]%
        {Ratner2019TrainingCM}
\bibfield{author}{\bibinfo{person}{Alexander Ratner}, \bibinfo{person}{Braden
  Hancock}, \bibinfo{person}{Jared Dunnmon}, \bibinfo{person}{Frederic Sala},
  \bibinfo{person}{Shreyash Pandey}, {and} \bibinfo{person}{Christopher
  R{\'e}}.} \bibinfo{year}{2019}\natexlab{}.
\newblock \showarticletitle{Training Complex Models with Multi-Task Weak
  Supervision}.
\newblock \bibinfo{journal}{\emph{Proceedings of the ... AAAI Conference on
  Artificial Intelligence. AAAI Conference on Artificial Intelligence}}
  \bibinfo{volume}{33} (\bibinfo{year}{2019}), \bibinfo{pages}{4763--4771}.
\newblock


\bibitem[\protect\citeauthoryear{Riordan, Horbach, Cahill, Zesch, and
  Lee}{Riordan et~al\mbox{.}}{2017}]%
        {Riordan2017InvestigatingNA}
\bibfield{author}{\bibinfo{person}{Brian Riordan}, \bibinfo{person}{Andrea
  Horbach}, \bibinfo{person}{Aoife Cahill}, \bibinfo{person}{Torsten Zesch},
  {and} \bibinfo{person}{Chong~Min Lee}.} \bibinfo{year}{2017}\natexlab{}.
\newblock \showarticletitle{Investigating neural architectures for short answer
  scoring}. In \bibinfo{booktitle}{\emph{BEA@EMNLP}}.
\newblock


\bibitem[\protect\citeauthoryear{Roy, Bhatt, and Narahari}{Roy
  et~al\mbox{.}}{2016}]%
        {Roy2016AnIT}
\bibfield{author}{\bibinfo{person}{Shourya Roy}, \bibinfo{person}{Himanshu~S.
  Bhatt}, {and} \bibinfo{person}{Y. Narahari}.}
  \bibinfo{year}{2016}\natexlab{}.
\newblock \showarticletitle{An Iterative Transfer Learning Based Ensemble
  Technique for Automatic Short Answer Grading}.
\newblock \bibinfo{journal}{\emph{ArXiv}}  \bibinfo{volume}{abs/1609.04909}
  (\bibinfo{year}{2016}).
\newblock


\bibitem[\protect\citeauthoryear{Saha, Dhamecha, Marvaniya, Foltz, Sindhgatta,
  and Sengupta}{Saha et~al\mbox{.}}{2019}]%
        {Saha2019JointML}
\bibfield{author}{\bibinfo{person}{Swarnadeep Saha}, \bibinfo{person}{Tejas~I.
  Dhamecha}, \bibinfo{person}{Smit Marvaniya}, \bibinfo{person}{Peter Foltz},
  \bibinfo{person}{Renuka Sindhgatta}, {and} \bibinfo{person}{Bikram
  Sengupta}.} \bibinfo{year}{2019}\natexlab{}.
\newblock \showarticletitle{Joint Multi-Domain Learning for Automatic Short
  Answer Grading}.
\newblock \bibinfo{journal}{\emph{ArXiv}}  \bibinfo{volume}{abs/1902.09183}
  (\bibinfo{year}{2019}).
\newblock


\bibitem[\protect\citeauthoryear{Saha, Dhamecha, Marvaniya, Sindhgatta, and
  Sengupta}{Saha et~al\mbox{.}}{2018}]%
        {Saha2018SentenceLO}
\bibfield{author}{\bibinfo{person}{Swarnadeep Saha}, \bibinfo{person}{Tejas~I.
  Dhamecha}, \bibinfo{person}{Smit Marvaniya}, \bibinfo{person}{Renuka
  Sindhgatta}, {and} \bibinfo{person}{Bikram Sengupta}.}
  \bibinfo{year}{2018}\natexlab{}.
\newblock \showarticletitle{Sentence Level or Token Level Features for
  Automatic Short Answer Grading?: Use Both}. In
  \bibinfo{booktitle}{\emph{Artificial Intelligence in Education}},
  \bibfield{editor}{\bibinfo{person}{Carolyn Penstein~Ros{\'e}},
  \bibinfo{person}{Roberto Mart{\'i}nez-Maldonado}, \bibinfo{person}{H.~Ulrich
  Hoppe}, \bibinfo{person}{Rose Luckin}, \bibinfo{person}{Manolis Mavrikis},
  \bibinfo{person}{Kaska Porayska-Pomsta}, \bibinfo{person}{Bruce McLaren},
  {and} \bibinfo{person}{Benedict du~Boulay}} (Eds.).
  \bibinfo{publisher}{Springer International Publishing},
  \bibinfo{address}{Cham}, \bibinfo{pages}{503--517}.
\newblock
\showISBNx{978-3-319-93843-1}


\bibitem[\protect\citeauthoryear{Sahu and Bhowmick}{Sahu and Bhowmick}{2020}]%
        {Sahu2020FeatureEA}
\bibfield{author}{\bibinfo{person}{A. Sahu} {and} \bibinfo{person}{P.~K.
  Bhowmick}.} \bibinfo{year}{2020}\natexlab{}.
\newblock \showarticletitle{Feature Engineering and Ensemble-Based Approach for
  Improving Automatic Short-Answer Grading Performance}.
\newblock \bibinfo{journal}{\emph{IEEE Transactions on Learning Technologies}}
  \bibinfo{volume}{13} (\bibinfo{year}{2020}), \bibinfo{pages}{77--90}.
\newblock


\bibitem[\protect\citeauthoryear{Sultan, Salazar, and Sumner}{Sultan
  et~al\mbox{.}}{2016}]%
        {Sultan2016FastAE}
\bibfield{author}{\bibinfo{person}{Md~Arafat Sultan},
  \bibinfo{person}{Cristobal Salazar}, {and} \bibinfo{person}{Tamara Sumner}.}
  \bibinfo{year}{2016}\natexlab{}.
\newblock \showarticletitle{Fast and Easy Short Answer Grading with High
  Accuracy}. In \bibinfo{booktitle}{\emph{Proceedings of the 2016 Conference of
  the North {A}merican Chapter of the Association for Computational
  Linguistics: Human Language Technologies}}. \bibinfo{publisher}{Association
  for Computational Linguistics}, \bibinfo{address}{San Diego, California},
  \bibinfo{pages}{1070--1075}.
\newblock
\urldef\tempurl%
\url{https://doi.org/10.18653/v1/N16-1123}
\showDOI{\tempurl}


\bibitem[\protect\citeauthoryear{Sung, Dhamecha, and Mukhi}{Sung
  et~al\mbox{.}}{2019a}]%
        {Sung2019ImprovingSA}
\bibfield{author}{\bibinfo{person}{Chul Sung}, \bibinfo{person}{Tejas~Indulal
  Dhamecha}, {and} \bibinfo{person}{Nirmal Mukhi}.}
  \bibinfo{year}{2019}\natexlab{a}.
\newblock \showarticletitle{Improving Short Answer Grading Using
  Transformer-Based Pre-training}. In \bibinfo{booktitle}{\emph{Artificial
  Intelligence in Education}}, \bibfield{editor}{\bibinfo{person}{Seiji
  Isotani}, \bibinfo{person}{Eva Mill{\'a}n}, \bibinfo{person}{Amy Ogan},
  \bibinfo{person}{Peter Hastings}, \bibinfo{person}{Bruce McLaren}, {and}
  \bibinfo{person}{Rose Luckin}} (Eds.). \bibinfo{publisher}{Springer
  International Publishing}, \bibinfo{address}{Cham},
  \bibinfo{pages}{469--481}.
\newblock
\showISBNx{978-3-030-23204-7}


\bibitem[\protect\citeauthoryear{Sung, Dhamecha, Saha, Ma, Reddy, and
  Arora}{Sung et~al\mbox{.}}{2019b}]%
        {Sung2019PreTrainingBO}
\bibfield{author}{\bibinfo{person}{Chul Sung}, \bibinfo{person}{Tejas~I.
  Dhamecha}, \bibinfo{person}{Swarnadeep Saha}, \bibinfo{person}{Tengfei Ma},
  \bibinfo{person}{V.~Pulla Reddy}, {and} \bibinfo{person}{Rishi Arora}.}
  \bibinfo{year}{2019}\natexlab{b}.
\newblock \showarticletitle{Pre-Training BERT on Domain Resources for Short
  Answer Grading}. In \bibinfo{booktitle}{\emph{EMNLP/IJCNLP}}.
\newblock


\bibitem[\protect\citeauthoryear{Sutskever, Vinyals, and Le}{Sutskever
  et~al\mbox{.}}{2014}]%
        {Sutskever14}
\bibfield{author}{\bibinfo{person}{Ilya Sutskever}, \bibinfo{person}{Oriol
  Vinyals}, {and} \bibinfo{person}{Quoc~V. Le}.}
  \bibinfo{year}{2014}\natexlab{}.
\newblock \showarticletitle{Sequence to Sequence Learning with Neural
  Networks}. In \bibinfo{booktitle}{\emph{Proceedings of the 27th International
  Conference on Neural Information Processing Systems - Volume 2}}
  \emph{(\bibinfo{series}{NIPS'14})}. \bibinfo{publisher}{MIT Press},
  \bibinfo{address}{Cambridge, MA, USA}, \bibinfo{pages}{3104–3112}.
\newblock


\bibitem[\protect\citeauthoryear{Suzen, Gorban, Levesley, and Mirkes}{Suzen
  et~al\mbox{.}}{2018}]%
        {Suzen2018}
\bibfield{author}{\bibinfo{person}{Neslihan Suzen},
  \bibinfo{person}{Alexander~N. Gorban}, \bibinfo{person}{Jeremy Levesley},
  {and} \bibinfo{person}{Eugenij~Moiseevich Mirkes}.}
  \bibinfo{year}{2018}\natexlab{}.
\newblock \showarticletitle{Automatic Short Answer Grading and Feedback Using
  Text Mining Methods}.
\newblock \bibinfo{journal}{\emph{ArXiv}}  \bibinfo{volume}{abs/1807.10543}
  (\bibinfo{year}{2018}).
\newblock


\bibitem[\protect\citeauthoryear{Tevet, Habib, Shwartz, and Berant}{Tevet
  et~al\mbox{.}}{2019}]%
        {Tevet2019}
\bibfield{author}{\bibinfo{person}{Guy Tevet}, \bibinfo{person}{Gavriel Habib},
  \bibinfo{person}{Vered Shwartz}, {and} \bibinfo{person}{Jonathan Berant}.}
  \bibinfo{year}{2019}\natexlab{}.
\newblock \showarticletitle{Evaluating Text {GAN}s as Language Models}. In
  \bibinfo{booktitle}{\emph{Proceedings of the 2019 Conference of the North
  {A}merican Chapter of the Association for Computational Linguistics: Human
  Language Technologies, Volume 1 (Long and Short Papers)}}.
  \bibinfo{publisher}{Association for Computational Linguistics},
  \bibinfo{address}{Minneapolis, Minnesota}, \bibinfo{pages}{2241--2247}.
\newblock
\urldef\tempurl%
\url{https://doi.org/10.18653/v1/N19-1233}
\showDOI{\tempurl}


\bibitem[\protect\citeauthoryear{Tsiakmaki, Kostopoulos, Kotsiantis, and
  Ragos}{Tsiakmaki et~al\mbox{.}}{2020}]%
        {Tsiakmaki2020}
\bibfield{author}{\bibinfo{person}{Maria Tsiakmaki}, \bibinfo{person}{Georgios
  Kostopoulos}, \bibinfo{person}{Sotiris Kotsiantis}, {and}
  \bibinfo{person}{Omiros Ragos}.} \bibinfo{year}{2020}\natexlab{}.
\newblock \showarticletitle{Transfer Learning from Deep Neural Networks for
  Predicting Student Performance}.
\newblock \bibinfo{journal}{\emph{Applied Sciences}} \bibinfo{volume}{10},
  \bibinfo{number}{6} (\bibinfo{year}{2020}).
\newblock
\showISSN{2076-3417}
\urldef\tempurl%
\url{https://doi.org/10.3390/app10062145}
\showDOI{\tempurl}


\bibitem[\protect\citeauthoryear{van~der Lee and van~den Bosch}{van~der Lee and
  van~den Bosch}{2017}]%
        {Lee2017ExploringLA}
\bibfield{author}{\bibinfo{person}{Chris van~der Lee} {and}
  \bibinfo{person}{Antal van~den Bosch}.} \bibinfo{year}{2017}\natexlab{}.
\newblock \showarticletitle{Exploring Lexical and Syntactic Features for
  Language Variety Identification}. In \bibinfo{booktitle}{\emph{Proceedings of
  the Fourth Workshop on {NLP} for Similar Languages, Varieties and Dialects
  ({V}ar{D}ial)}}. \bibinfo{publisher}{Association for Computational
  Linguistics}, \bibinfo{address}{Valencia, Spain}, \bibinfo{pages}{190--199}.
\newblock
\urldef\tempurl%
\url{https://doi.org/10.18653/v1/W17-1224}
\showDOI{\tempurl}


\bibitem[\protect\citeauthoryear{Vaswani, Shazeer, Parmar, Uszkoreit, Jones,
  Gomez, Kaiser, and Polosukhin}{Vaswani et~al\mbox{.}}{2017}]%
        {Vaswani2017AttentionIA}
\bibfield{author}{\bibinfo{person}{Ashish Vaswani}, \bibinfo{person}{Noam
  Shazeer}, \bibinfo{person}{Niki Parmar}, \bibinfo{person}{Jakob Uszkoreit},
  \bibinfo{person}{Llion Jones}, \bibinfo{person}{Aidan~N Gomez},
  \bibinfo{person}{\L~ukasz Kaiser}, {and} \bibinfo{person}{Illia Polosukhin}.}
  \bibinfo{year}{2017}\natexlab{}.
\newblock \showarticletitle{Attention is All you Need}.
\newblock   \bibinfo{volume}{30} (\bibinfo{year}{2017}).
\newblock
\urldef\tempurl%
\url{https://proceedings.neurips.cc/paper/2017/file/3f5ee243547dee91fbd053c1c4a845aa-Paper.pdf}
\showURL{%
\tempurl}


\bibitem[\protect\citeauthoryear{Wang, Inoue, Ouchi, Mizumoto, and Inui}{Wang
  et~al\mbox{.}}{2019}]%
        {Wang2019InjectRI}
\bibfield{author}{\bibinfo{person}{Tianqi Wang}, \bibinfo{person}{Naoya Inoue},
  \bibinfo{person}{Hiroki Ouchi}, \bibinfo{person}{Tomoya Mizumoto}, {and}
  \bibinfo{person}{Kentaro Inui}.} \bibinfo{year}{2019}\natexlab{}.
\newblock \showarticletitle{Inject Rubrics into Short Answer Grading System}.
  In \bibinfo{booktitle}{\emph{Proceedings of the 2nd Workshop on Deep Learning
  Approaches for Low-Resource NLP (DeepLo 2019)}}.
  \bibinfo{publisher}{Association for Computational Linguistics},
  \bibinfo{address}{Hong Kong, China}, \bibinfo{pages}{175--182}.
\newblock
\urldef\tempurl%
\url{https://doi.org/10.18653/v1/D19-6119}
\showDOI{\tempurl}


\bibitem[\protect\citeauthoryear{Wang, Mizumoto, Inoue, and Inui}{Wang
  et~al\mbox{.}}{2018}]%
        {Wang2018IdentifyingCI}
\bibfield{author}{\bibinfo{person}{Tianqi Wang}, \bibinfo{person}{Tomoya
  Mizumoto}, \bibinfo{person}{Naoya Inoue}, {and} \bibinfo{person}{Kentaro
  Inui}.} \bibinfo{year}{2018}\natexlab{}.
\newblock \showarticletitle{Identifying Current Issues in Short Answer
  Grading}.
\newblock


\bibitem[\protect\citeauthoryear{Wang, Wang, Qin, Packer, Li, Chen, Beutel, and
  Chi}{Wang et~al\mbox{.}}{2020}]%
        {CATGEN}
\bibfield{author}{\bibinfo{person}{Tianlu Wang}, \bibinfo{person}{Xuezhi Wang},
  \bibinfo{person}{Yao Qin}, \bibinfo{person}{Ben Packer},
  \bibinfo{person}{Kang Li}, \bibinfo{person}{Jilin Chen},
  \bibinfo{person}{Alex Beutel}, {and} \bibinfo{person}{Ed Chi}.}
  \bibinfo{year}{2020}\natexlab{}.
\newblock \showarticletitle{{CAT}-Gen: Improving Robustness in {NLP} Models via
  Controlled Adversarial Text Generation}. In
  \bibinfo{booktitle}{\emph{Proceedings of the 2020 Conference on Empirical
  Methods in Natural Language Processing (EMNLP)}}.
  \bibinfo{publisher}{Association for Computational Linguistics},
  \bibinfo{address}{Online}, \bibinfo{pages}{5141--5146}.
\newblock
\urldef\tempurl%
\url{https://doi.org/10.18653/v1/2020.emnlp-main.417}
\showDOI{\tempurl}


\bibitem[\protect\citeauthoryear{Wang, Wang, Wang, Wang, and Ye}{Wang
  et~al\mbox{.}}{2021}]%
        {wang2021robust}
\bibfield{author}{\bibinfo{person}{Wenqi Wang}, \bibinfo{person}{Run Wang},
  \bibinfo{person}{Lina Wang}, \bibinfo{person}{Zhibo Wang}, {and}
  \bibinfo{person}{Aoshuang Ye}.} \bibinfo{year}{2021}\natexlab{}.
\newblock \bibinfo{title}{Towards a Robust Deep Neural Network in Texts: A
  Survey}.
\newblock
\newblock
\showeprint[arxiv]{cs.CL/1902.07285}


\bibitem[\protect\citeauthoryear{Wei and Zou}{Wei and Zou}{2019}]%
        {Wei2019}
\bibfield{author}{\bibinfo{person}{Jason Wei} {and} \bibinfo{person}{Kai Zou}.}
  \bibinfo{year}{2019}\natexlab{}.
\newblock \showarticletitle{{EDA}: Easy Data Augmentation Techniques for
  Boosting Performance on Text Classification Tasks}. In
  \bibinfo{booktitle}{\emph{Proceedings of the 2019 Conference on Empirical
  Methods in Natural Language Processing and the 9th International Joint
  Conference on Natural Language Processing (EMNLP-IJCNLP)}}.
  \bibinfo{address}{Hong Kong, China}, \bibinfo{pages}{6382--6388}.
\newblock
\urldef\tempurl%
\url{https://doi.org/10.18653/v1/D19-1670}
\showDOI{\tempurl}


\bibitem[\protect\citeauthoryear{Yang, Dai, Yang, Carbonell, Salakhutdinov, and
  Le}{Yang et~al\mbox{.}}{2019}]%
        {Yang2019XLNetGA}
\bibfield{author}{\bibinfo{person}{Zhilin Yang}, \bibinfo{person}{Zihang Dai},
  \bibinfo{person}{Yiming Yang}, \bibinfo{person}{Jaime Carbonell},
  \bibinfo{person}{Russ~R Salakhutdinov}, {and} \bibinfo{person}{Quoc~V Le}.}
  \bibinfo{year}{2019}\natexlab{}.
\newblock \showarticletitle{XLNet: Generalized Autoregressive Pretraining for
  Language Understanding}. In \bibinfo{booktitle}{\emph{Advances in Neural
  Information Processing Systems}},
  \bibfield{editor}{\bibinfo{person}{H.~Wallach},
  \bibinfo{person}{H.~Larochelle}, \bibinfo{person}{A.~Beygelzimer},
  \bibinfo{person}{F.~d\textquotesingle Alch\'{e}-Buc},
  \bibinfo{person}{E.~Fox}, {and} \bibinfo{person}{R.~Garnett}} (Eds.),
  Vol.~\bibinfo{volume}{32}. \bibinfo{publisher}{Curran Associates, Inc.}
\newblock
\urldef\tempurl%
\url{https://proceedings.neurips.cc/paper/2019/file/dc6a7e655d7e5840e66733e9ee67cc69-Paper.pdf}
\showURL{%
\tempurl}


\bibitem[\protect\citeauthoryear{Yoon, Denton, Hoang, and Rush}{Yoon
  et~al\mbox{.}}{2017}]%
        {KimDHR17}
\bibfield{author}{\bibinfo{person}{Yoon}, \bibinfo{person}{Carl Denton},
  \bibinfo{person}{Luong Hoang}, {and} \bibinfo{person}{Alexander~M. Rush}.}
  \bibinfo{year}{2017}\natexlab{}.
\newblock \showarticletitle{Structured Attention Networks}.
\newblock \bibinfo{journal}{\emph{CoRR}}  \bibinfo{volume}{abs/1702.00887}
  (\bibinfo{year}{2017}).
\newblock
\showeprint[arxiv]{1702.00887}


\bibitem[\protect\citeauthoryear{Young, Hazarika, Poria, and Cambria}{Young
  et~al\mbox{.}}{2018}]%
        {Young2018}
\bibfield{author}{\bibinfo{person}{Tom Young}, \bibinfo{person}{Devamanyu
  Hazarika}, \bibinfo{person}{Soujanya Poria}, {and} \bibinfo{person}{Erik
  Cambria}.} \bibinfo{year}{2018}\natexlab{}.
\newblock \showarticletitle{Recent Trends in Deep Learning Based Natural
  Language Processing [Review Article]}.
\newblock \bibinfo{journal}{\emph{IEEE Computational Intelligence Magazine}}
  \bibinfo{volume}{13}, \bibinfo{number}{3} (\bibinfo{year}{2018}),
  \bibinfo{pages}{55--75}.
\newblock
\urldef\tempurl%
\url{https://doi.org/10.1109/MCI.2018.2840738}
\showDOI{\tempurl}


\bibitem[\protect\citeauthoryear{Zhang, Huang, Yang, Yu, and Zhuang}{Zhang
  et~al\mbox{.}}{2019}]%
        {Zhang2019AnAS}
\bibfield{author}{\bibinfo{person}{Lishan Zhang}, \bibinfo{person}{Yuwei
  Huang}, \bibinfo{person}{Xi Yang}, \bibinfo{person}{Shengquan Yu}, {and}
  \bibinfo{person}{Fuzhen Zhuang}.} \bibinfo{year}{2019}\natexlab{}.
\newblock \showarticletitle{An automatic short-answer grading model for
  semi-open-ended questions}.
\newblock \bibinfo{journal}{\emph{Interactive Learning Environments}}
  \bibinfo{volume}{0}, \bibinfo{number}{0} (\bibinfo{year}{2019}),
  \bibinfo{pages}{1--14}.
\newblock
\urldef\tempurl%
\url{https://doi.org/10.1080/10494820.2019.1648300}
\showDOI{\tempurl}


\bibitem[\protect\citeauthoryear{Zhang, Lin, and Chi}{Zhang
  et~al\mbox{.}}{2020a}]%
        {Zhang2020GoingDA}
\bibfield{author}{\bibinfo{person}{Yuan Zhang}, \bibinfo{person}{Chen Lin},
  {and} \bibinfo{person}{Min Chi}.} \bibinfo{year}{2020}\natexlab{a}.
\newblock \showarticletitle{Going deeper: Automatic short-answer grading by
  combining student and question models}.
\newblock \bibinfo{journal}{\emph{User Modeling and User-Adapted Interaction}}
  \bibinfo{volume}{30}, \bibinfo{number}{1} (\bibinfo{date}{01 Mar}
  \bibinfo{year}{2020}), \bibinfo{pages}{51--80}.
\newblock
\showISSN{1573-1391}
\urldef\tempurl%
\url{https://doi.org/10.1007/s11257-019-09251-6}
\showDOI{\tempurl}


\bibitem[\protect\citeauthoryear{Zhang, Wu, Zhao, Li, Zhang, Zhou, and
  Zhou}{Zhang et~al\mbox{.}}{2020b}]%
        {Zhang2019SemanticsawareBF}
\bibfield{author}{\bibinfo{person}{Zhuosheng Zhang}, \bibinfo{person}{Yuwei
  Wu}, \bibinfo{person}{Hai Zhao}, \bibinfo{person}{Zuchao Li},
  \bibinfo{person}{Shuailiang Zhang}, \bibinfo{person}{Xi Zhou}, {and}
  \bibinfo{person}{Xiang Zhou}.} \bibinfo{year}{2020}\natexlab{b}.
\newblock \showarticletitle{Semantics-Aware BERT for Language Understanding}.
\newblock \bibinfo{journal}{\emph{Proceedings of the AAAI Conference on
  Artificial Intelligence}} \bibinfo{volume}{34}, \bibinfo{number}{05}
  (\bibinfo{date}{Apr.} \bibinfo{year}{2020}), \bibinfo{pages}{9628--9635}.
\newblock
\urldef\tempurl%
\url{https://doi.org/10.1609/aaai.v34i05.6510}
\showDOI{\tempurl}


\end{thebibliography}
\end{document}